\documentclass[letterpaper, 10 pt, conference]{ieeeconf}  

\IEEEoverridecommandlockouts                              
\usepackage[font=small,labelfont=bf]{caption}
\usepackage{hyperref}
\usepackage{amsmath,amsfonts,amssymb}
\usepackage{graphicx}
\usepackage{prettyref}
\usepackage{mathrsfs}
\usepackage{wrapfig}
\usepackage{MnSymbol}
\usepackage{multirow}
\usepackage{xcolor}
\usepackage{booktabs}
\usepackage{algorithm,algpseudocode}
\usepackage{subfig}
\algnewcommand{\LineComment}[1]{\State \(\triangleright\) #1}

\newrefformat{Fig}{Figure~\ref{#1}}
\newrefformat{fig}{Figure~\ref{#1}}
\newrefformat{par}{Section~\ref{#1}}
\newrefformat{appen}{Appendix~\ref{#1}}
\newrefformat{sec}{Section~\ref{#1}}
\newrefformat{sub}{Section~\ref{#1}}
\newrefformat{table}{Table~\ref{#1}}
\newrefformat{alg}{Algorithm~\ref{#1}}
\newrefformat{Alg}{Algorithm~\ref{#1}}
\newrefformat{Def}{Definition~\ref{#1}}
\newrefformat{Thm}{Theorem~\ref{#1}}
\newrefformat{step}{Step~\ref{#1}}
\newrefformat{ln}{Line~\ref{#1}}
\newrefformat{eq}{Equation~\ref{#1}}
\newrefformat{pb}{Problem~\ref{#1}}
\newrefformat{it}{Item~\ref{#1}}
\newrefformat{te}{Term~\ref{#1}}
\def\Eqref Eq:#1:{\eqref{eq:#1}}
\newrefformat{Eq}{Equation~\Eqref#1:}

\newcommand{\E}[1]{\mathbf{#1}}
\newcommand{\TE}[1]{\textbf{#1}}

\newcommand{\FDM}[2]{\frac{D{#1}}{D{#2}}}

\newcommand{\argmin}[1]{\underset{#1}{\E{argmin}}}

\usepackage{tikz}
\usetikzlibrary[arrows,backgrounds,decorations.pathmorphing,positioning,fit,petri]

\newcommand{\changed}[1]{\textcolor{black}{#1}}
\newenvironment{changedBlk}{\color{black}}{}
\newcommand{\COMM}[2]{}{}
\newcommand{\FINE}[1]{}
\newcommand{\TDS}{{TRANSFER+FOLLOW}}
\newcommand{\TSDS}{{TRANSFER+ZERO}}
\newcommand{\LOP}{{liquid outflow curve}}
\newcommand{\MTP}{{mean trajectory prior}}
\newcommand{\CC}{{computational complexity}}

\title{\LARGE \bf Feedback Motion Planning for Liquid \changed{Pouring} \\Using Supervised Learning}
\author{Zherong Pan$^{1}$ and Dinesh Manocha$^{1}$  \\
\url{http://gamma.cs.unc.edu/RLFluid}
\thanks{$^{1}$ Department of Computer Science, the University of North Carolina at Chapel Hill {\tt\small \{zherong,dm\}@cs.unc.edu}}%
}

\begin{document}

\maketitle
\thispagestyle{empty}
\pagestyle{empty}

\begin{abstract}
We present a novel motion planning algorithm for pouring a liquid body from a source to a target container. \changed{Our approach uses a receding-horizon optimization strategy that takes into account liquid dynamics and various other constraints. In order to handle liquid dynamics without costly fluid simulations, we use a neural network to infer a set of key liquid-related parameters from the observation of current liquid configuration. To train the neural network, we generate a dataset of successful pouring examples using stochastic optimization in a problem-specific search space. These parameters are then used in the objective function for trajectory optimization. Our feedback motion planner achieves real-time performance, and we observe a high success rate in our simulated 2D and 3D liquid pouring benchmarks.}
\end{abstract}

\section{INTRODUCTION}\label{sec:intro}
Robotic manipulation of non-rigid objects such as fluids,  elastic bodies, and strings is a challenging problem that arises in different applications. \changed{In this paper, we address the problem of pouring liquids, where the goal is to use a robot to pour a liquid body from a source to a target container. These tasks arise when industrial robots are used for painting, cleaning, or dispensing lubricants. Other applications include the use of service robots for cooking, cleaning, or feeding.}

A key issue in such motion planning algorithms is to satisfy the liquid dynamics constraints. The liquid body can have a complex topology and undergo large deformations. The underlying solvers tend to use a large number of particles (tens of thousands) to model their motion. This results in a very high dimensional configuration space of the liquid body and makes it hard to directly use sampling-based motion planning algorithms. Other techniques based on an optimization-based planner \cite{STOMP:2011,schulman2014motion} may not work well because the free-surface of a liquid body introduces non-smooth changes.

Prior planning algorithms for fluid manipulation are either based on demonstration and learning methods, or use dynamics constraints. \changed{The demonstration-based methods use example trajectories and ignore all physical constraints so that they may not generalize to new scenarios. On the other hand, methods using reinforcement learning \cite{yamaguchi2016neural,yamaguchi2015differential} can take physics constraints into consideration but require a problem specific training dataset for each manipulation task. Other techniques use trajectory optimization, which takes into account a full-featured liquid dynamics model \cite{kuriyama2008trajectory,pan2016robot}, but these techniques have a very high computational overhead. This problem can be alleviated using reduced or simplified dynamics models \cite{kunze2011simulation,tzamtzi2008robustness,pan2016simple}, where many liquid dynamics constraints are ignored, and these models are combined with open-loop planners. However, open-loop methods suffer from simulation bias.}

\begin{figure}
\begin{center}
\includegraphics[trim=0mm 0mm 0mm 0mm, clip, width=.99\linewidth]{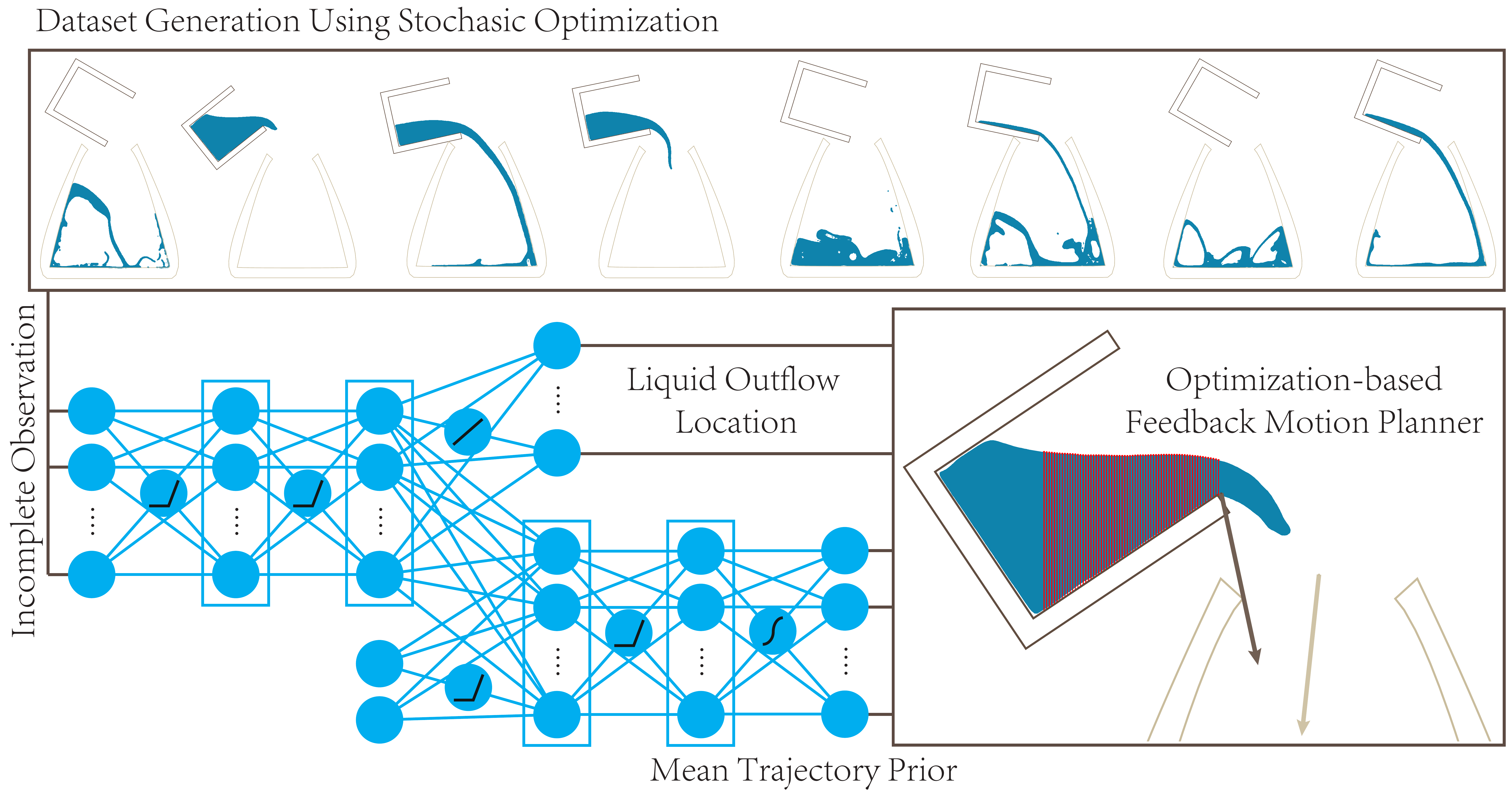}
\put (-202,88 ) {(a)}
\put (-210,20 ) {(b)}
\put (-70 ,20 ) {(c)}
\end{center}
\vspace{-5px}
\caption{\label{Fig:teaser} An illustration of our feedback motion planning framework. From the training dataset found by stochastic optimization (a), we train a neural network that predicts liquid-related parameters: \LOP{} and the \MTP{} (b). Our online planner then computes the source container trajectory from an incomplete observation (c), e.g., the liquid height field in red and the moving speed of the source/target containers (the two arrows).}
\vspace{-15px}
\end{figure}

\noindent{\bf Main Results:} \begin{changedBlk} 
We present a feedback motion planning algorithm to control the fluid flow for liquid pouring subject to the dynamics constraints. In order to handle the  high dimensionality of fluids, we present a learning-based approach to predict the liquid configurations based on low-dimensional features. This is combined with a receding-horizon trajectory optimization method for online planning. The main novel components of our approach include:

\begin{itemize}
\item A supervised learning algorithm that enables predicting the state of a high-dimensional liquid body using low-dimensional features. We also describe an efficient strategy to train the neural network.
\item A feedback motion planning that solves a spacetime optimization problem using the receding horizon strategy. The objective function is guided by the neural network, and this significantly increase the efficiency of our planner, as we do not need to perform the costly 3D fluid simulation step at each timestep.
\end{itemize}

Our neural network is trained using a large amount of successful pouring trajectories. This dataset is automatically generated offline using a large number of random configurations of liquid pouring problems. These configurations are generated by changing the relative positions of two containers, the amount of liquid in the source container, and the speed of the target container movement. As a result, the trained neural network is robust to environmental changes. We compute the optimal liquid pouring trajectory for each configuration using stochastic optimizations, with the help of a fully-featured liquid simulator.

We have evaluated our algorithm on many new and challenging scenarios that are quite different from the training dataset. In practice, our learning-based planner achieves almost the same success rate as was obtained using the groundtruth trajectories on our training dataset. Moreover, we have also tested our planner in new and different environments, including 3D workspaces, liquids with different physical characteristics (e.g. varying the viscosity), and different container shapes. Our results demonstrate that, the neural networking can be very robust when used with a optimization-based motion planner, although our training dataset uses a single low-viscous liquid material and a rectangular container shape. Furthermore, the online algorithm is very fast and takes less than $10$ milliseconds on a single core of Intel CPU to plan the motion amongst dynamic obstacles.
\end{changedBlk}

\section{ASSUMPTIONS AND PROBLEM FORMULATION}
In this work, we restrict ourselves to a simulated environment, where the pouring problem is formulated in \prettyref{sec:problem}. Our algorithm is composed of three components (see \prettyref{Fig:teaser}). 

In the preprocessing stage, we generated a large set of random liquid pouring problems. For each problem, we find a successful pouring trajectory using stochastic optimization (see \prettyref{sec:learning}). We then extract a low dimensional feature of the liquid body and train a 4-layered neural network to predict a set of key parameters in pouring (\LOP{} and \MTP{}). We have not yet applied our framework in a real-life robotic system and currently only assume that the features can be acquired from the sensing data. However, we propose two kinds of possible liquid features to inspire future work on a real-life robotic system. Other features can also be easily used with our method.

During the online stage, the predicted parameters are used in the objective function of a receding-horizon optimization-based motion planner (see \prettyref{sec:framework}). The planner determines the configuration of the source container at the next timestep by minimizing an objective function that encourages successful pouring and also takes into account various other constraints including collision avoidance and trajectory smoothness. 

\section{RELATED WORKS}\label{sec:related}
Our approach builds on three areas of prior work: motion planning, planning for dynamic objects, and reinforcement learning. 

\subsection{GENERAL MOTION PLANNING}
A motion planning algorithm searches for a trajectory that satisfies a set of constraints (collision-free, smoothness), which may also be optimal under a given quality measure. Many early motion planners such as \cite{lavalle1998rapidly} and its descendants \cite{GRVO:2009,stilman2007task,berenson2009manipulation} consider only collision-free constraints. Unlike these methods, which tend to compute a trajectory by sampling in the space of possible trajectories, optimization-based motion planners such as \cite{schulman2014motion,STOMP:2011,Park:2012:ICAPS} can easily take into account other constraints, such as dynamics, smoothness, etc. Many of these approaches formulate the problem as a spacetime continuous optimization. Such optimization methods have also been used for liquid transfer \cite{pan2016robot,pan2016simple} based on simplified dynamics when limited to static environments.

There is considerable work on feedback motion planning that uses refinement schemes based on feedback control laws. This can be performed using replanning \cite{ferguson2006replanning,hauser2012responsiveness,koenig2005fast} or by formulating the problem as a Markov Decision Process \cite{Kurniawati08sarsop:efficient}. These ideas have been applied to high dimensional continuous systems such as humanoid robots \cite{levine2014learning,park2014high}. In this work, we present such a feedback motion planning algorithm for liquid transfer.

\subsection{PLANNING FOR DYNAMIC OBJECTS}
The extension of conventional motion planning algorithms to the manipulation of non-rigid objects has been addressed in the context of virtual suturing \cite{schulman2016learning}, cloth folding \cite{li2015folding}, and surgical simulation \cite{Chentanez:2009:ISN}. It can be challenging to deal with non-rigid objects with high-dimensional configuration spaces. This is especially the case with liquid manipulation tasks, where the dimension can be as high as several million (see \cite{pan2016robot} for a detailed discussion). For certain types of fluids such as smoke and fire, optimization-based motion planning can be adapted to solve the problem by exploiting the special structure of the resulting fluid simulator \cite{treuille2003keyframe,DBLP:journals/corr/PanM16}. However, it is non-trivial to extend these methods to control liquid bodies with non-smooth, rapidly-changing free surfaces. Moreover, prior methods are designed for offline applications and computationally very costly. Previous work \cite{pan2016simple} reduced the computational cost by using a much simplified liquid model, dependent on just two variables.

\subsection{IMITATION OR REINFORCEMENT LEARNING}
\changed{Reinforcement and imitation learning have been shown to be effective in terms of controlling high dimensional dynamic systems, e.g., a humanoid robot \cite{levine2014learning,2016-TOG-deepRL}. Recently, imitation learning has been used to perform liquid manipulation using example container trajectories from a human demonstrator \cite{bowen2014closed,yamaguchi2014learning,brandi2014generalizing}. However, the learning framework in this work does not take fluid dynamics constraints into account. Moreover, trajectories of liquid body shapes from real-life experiments have to be captured and digitized to construct the dataset, which is challenging in and of itself (see \cite{wang2009physically,gregson2012stochastic}). More recently, reinforcement learning has also been used to learn pouring of granular materials in \cite{yamaguchi2015differential,yamaguchi2016neural}. Our methods differ from these methods in that we only use supervised learning, but we combine it with trajectory optimization to enhance the robustness of our motion planner.}

\begin{changedBlk}
\section{OVERVIEW}\label{sec:problem}
In this section, we first introduce our formulation of the liquid pouring problem and then outline our motion planning algorithm.

\subsection{PROBLEM FORMULATION}
In each problem, we consider a source container denoted by rigid body $\E{S}$, a target container denoted by $\E{T}$, and the liquid body denoted by $\E{L}$. Without ambiguity, we reuse these symbols to denote their spacetime trajectory $\E{S}(t)$, $\E{T}(t)$, and $\E{L}(t)$, where $t$ is the time index. Among these three trajectories, the liquid body trajectory $\E{L}(t)$ is constrained by the Navier-Stokes equation, the governing PDE of the liquid body. Therefore, the configuration of $\E{L}$ evolves as $\E{L}(t+\Delta t)=f(\E{S}(t),\E{T}(t),\E{L}(t))$, taking $\E{S}(t),\E{T}(t)$ as boundary conditions. Here $f$ is a time-integration of the Navier-Stokes equation over timestep size $\Delta t$.

In practice, we define a discrete version of the problem using a particle-based spatial discretization scheme, the finite-difference temporal discretization scheme, and the same particle-based fluid simulator as \cite{pan2016robot}. As a result, all three bodies $\E{S}, \E{T}$, and $\E{L}$ become a set of particles as illustrated in \prettyref{Fig:discrete} (a). Their corresponding trajectories are sampled uniformly in time with timestep size $\Delta t$. In this setting, the time-integration function $f$ can be approximately evaluated using a discrete version of the Navier-Stokes equation detailed in \prettyref{appen:sph}.

In our problem setting, the target container trajectory $\E{T}(t)$ is given. The goal of our feedback motion planner is then to determine the source container trajectory $\E{S}(t)$, so that the induced liquid body trajectory $\E{L}(t)$ will have the liquid body end up inside the target container $\E{T}$.

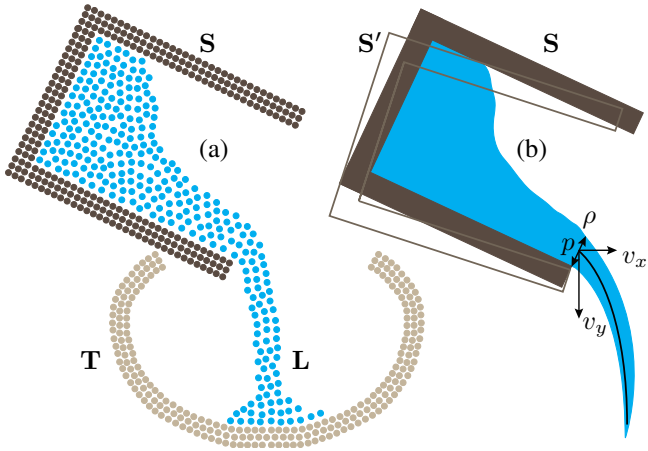
\begin{figure}
\begin{center}
\includegraphics[trim=0mm 0mm 0mm 0mm, clip, width=.99\linewidth]{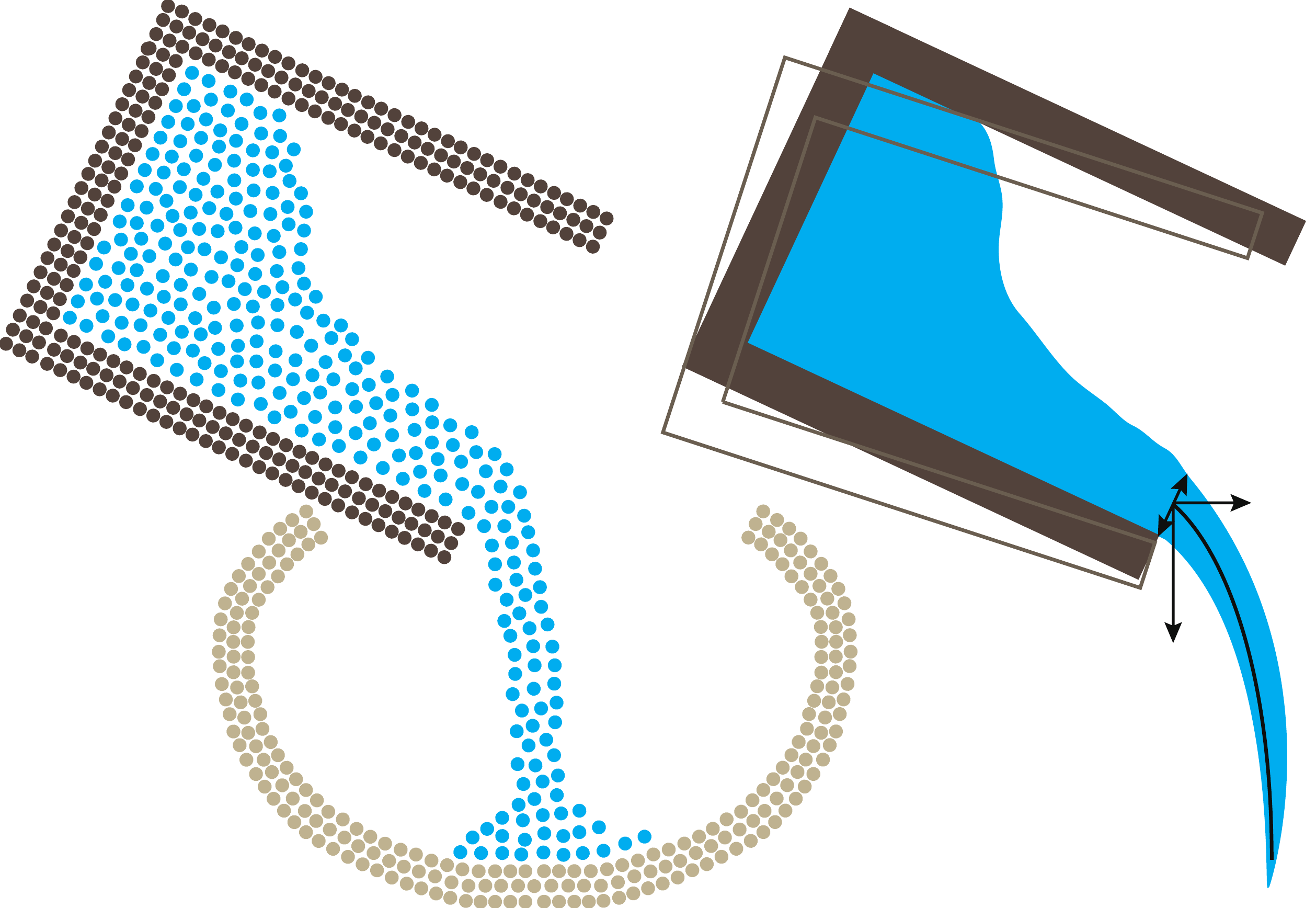}
\put (-170,150 ) {$\E{S}$}
\put (-215,30  ) {$\E{T}$}
\put (-135,30  ) {$\E{L}$}
\put (-170,110 ) {(a)}
\put (-40 ,150 ) {$\E{S}$}
\put (-110,150 ) {$\E{S}'$}
\put (-33 ,75  ) {$p$}
\put (-25 ,85  ) {$\rho$}
\put (-10 ,70  ) {$v_x$}
\put (-25 ,45  ) {$v_y$}
\put (-50 ,110 ) {(b)}
\end{center}
\caption{\label{Fig:discrete} \changed{(a): An illustration of our problem setting. We consider a source container $\E{S}$ in dark gray, a target container $\E{T}$ in light gray, and finally the liquid body $\E{L}$ in blue. We use a particle-based spatial discretization for both the rigid and liquid bodies. (b): The set of learned parameters used for formulating $C_o$: The \MTP{} $\E{S}'$, the parameters of \LOP{}: $p,v_x,v_y$, and the outflow flux $\rho$.}}
\vspace{-10px}
\end{figure}
\end{changedBlk}

\begin{changedBlk}
\section{FEEDBACK MOTION PLANNING}\label{sec:framework}
In order to find an entire trajectory $\E{S}(t)$, our motion planner iteratively solves the following spacetime optimization problem:
\begin{eqnarray}
\label{eq:Opt}
&&\argmin{\E{S}(t+\Delta t),\cdots,\E{S}(t+K\Delta t)}\sum_{1\leq k\leq K}C(\E{S}(t+k\Delta t))  \\
&&C(\E{S}(t))\triangleq C_{l}(\E{S}(t))+C_{r}(\E{S}(t))+C_{o}(\E{S}(t))\nonumber,
\end{eqnarray}
over a horizon of $K\Delta t$ and then only adopts the control $\E{S}(t+\Delta t)$, thus using the receding horizon strategy. The objective function $C$ involves three terms: The first term $C_l$ encourages the liquid body to fall inside $\E{T}$, $C_r$ is a regularization term that prefers smooth trajectories, and finally $C_o$ penalizes any collisions between $\E{S}$ and $\E{T}$ or any other obstacles in the environment. 

We can use the same $C_o$ as \cite{Park:2012:ICAPS,pan2016robot} for collision avoidance. Specifically, we define $C_o$ as:
\begin{eqnarray*}
C_o(\E{S}(t))=\|D(\E{S}(t))\|^2+\|\dot{D}(\E{S}(t))\|^2,
\end{eqnarray*}
where $D$ is the maximal penetration depth between $\E{S}$ and any obstacle. To accelerate collision detection, each rigid object is approximated using a set of spheres. Note that we penalize both the $D$ and $\dot{D}$ to encourage smooth movements when $\E{S}$ is in the vicinity of boundaries. The regularization term $\E{C}_r$ is the Laplace of the trajectory:
\begin{eqnarray*}
C_r(\E{S}(t))=\|\E{S}(t+\Delta t)-2\E{S}(t)+\E{S}(t-\Delta t)\|^2.
\end{eqnarray*}

In order to define $C_o$ such that it encourages successful liquid pouring, a naive and straightforward formulation would be to reconstruct $\E{L}(t)$ from $\E{S}(t),\E{T}(t)$ using liquid simulation function $f$, then to measure the amount of liquid that falls outside the target container and makes $C_o$ proportional to this amount. However, this formulation has two clear drawbacks: First, reconstructing $\E{L}(t)$ using liquid simulation function $f$ is computationally very costly, making it impossible for the motion planner to respond in real-time. Second, the function $f$ involves a lot of non-smooth operators, making it difficult for numerical optimizers to find a good local minimum of \prettyref{eq:Opt}.

Our solution is to base $C_o$ on a set of low-dimensional parameters that can be inferred from an observation of the current configuration of the liquid body $\E{L}(t)$ without resorting to its predicted future configurations. This idea is similar to the temporal decomposition method \cite{yamaguchi2015differential}. We assume that two kinds of parameters are crucial to the task of liquid pouring. First, there should be a pattern in most human pouring examples, in which the source container $\E{S}$ moves closer to the target container $\E{T}$, and at the same time increases its turning angle so that liquid can flow out. This common pattern is encoded as \MTP{} $\E{S}'(t)$. In addition, after liquid leaves the source container $\E{S}$, the flow can be well approximated as a quadratic curve, as shown in \cite{pan2016simple}. This quadratic curve could be used to guide the motion planner so that the curve is centered around the target container opening, thus avoiding spillage. In a 2D workspace, this curve is characterized by its starting point $p$ on $\E{S}$ and its leaving velocity $v_x,v_y$, as illustrated in \prettyref{Fig:discrete} (b). In summary, the required liquid-related parameters are $(\E{S}'(t),p,v_x,v_y)$, and we then define $C_o$ as:
\begin{eqnarray*}
\E{C}_o(\E{S}(t))&=&\|dist(\E{T}(t),p,v_x,v_y)\|^2\E{max}(\rho,0)+    \\
&&\|\E{S}(t)-\E{S}'(t)\|^2,
\end{eqnarray*}
where the first term measures the distance between \LOP{} and the center of target container opening, while the second term measures the discrepancy between the current source container configuration and \MTP{}, leading to a successful pouring. Note that we added a weighting $\E{max}(\rho,0)$ where $\rho$ is the predicted outflow flux. Therefore, we only apply the first term if liquid is flowing out. This $\rho$ is added to the required set of parameters. These parameters $(\E{S}'(t),p,v_x,v_y,\rho)$ are predicted efficiently using supervised learning introduced in \prettyref{sec:learning}.
\end{changedBlk}

\begin{changedBlk}
\section{SUPERVISED LEARNING}\label{sec:learning}
In this section, we present our preprocessing algorithm that uses supervised learning to predict the liquid-related parameters. 

As part of the objective function, the learning algorithm need to infer the parameters $(\E{S}'(t),p,v_x,v_y,\rho)$ from the observations $O(\E{L}(t))$ of liquid body and the configurations of the two rigid bodies $\E{S}(t),\E{T}(t)$. The inference is accomplished with a neural network trained from a set of successful liquid pouring trajectories. See \prettyref{Fig:DNN} for a specification of our neural network. 

The problem of inferring \MTP{} $\E{S}'(t)$ from the current configuration makes our neural network a control policy representation that can be optimized using reinforcement learning as in \cite{yamaguchi2015differential}. Optimized this way, the policy will predict $\E{S}'(t)$ that can be realized directly. Instead, only supervised learning is used in our work, so that the learned policy may suffer from simulation-bias issues and have limited ability to generalize to new environments. However, we can solve this problem by plugging the predicted $\E{S}'(t)$ into \prettyref{eq:Opt} to further adjust the predicted result.
\begin{figure}
\begin{center}
\includegraphics[trim=0mm 0mm 0mm 0mm, clip, width=.99\linewidth]{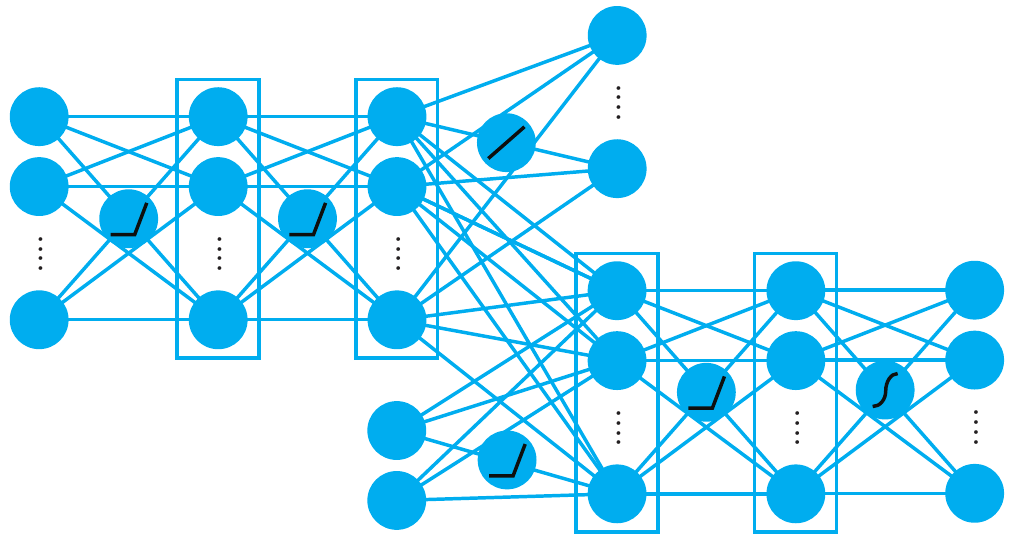}
\put (-240,35  ) {$O(\E{L}(t))$}
\put (-176,23  ) {$\E{S}(t)$}
\put (-178,6   ) {$\E{T}(t)$}
\put (-80 ,100 ) {$(p,v_x,v_y,\rho)$}
\put (-20 ,72  ) {$\E{S}'(t)$}
\end{center}
\caption{\label{Fig:DNN} \changed{Our 4-layer neural network structure for parameter estimation. The input is an observation of the current liquid configuration $O(L(t))$ and other rigid body configurations $\E{S}(t),\E{T}(t)$ where we assume the rigid body configurations are fully observable. We use ReLU activation function for all internal layers, sigmoid activation function for the output layer of \MTP{} $\E{S}'(t)$, and linear function for the output layer of \LOP{} parameters. Each of the four hidden layers has 32 hidden units.}}
\vspace{-15px}
\end{figure}
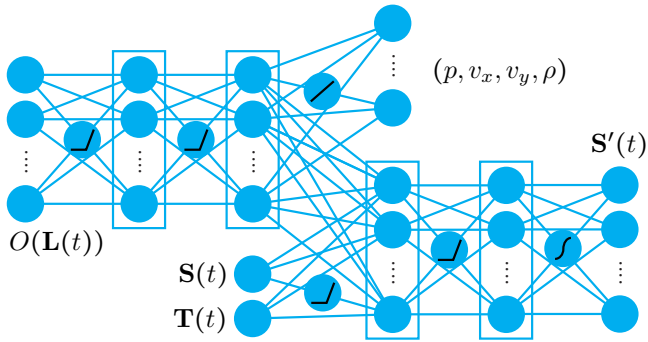

The observation function $O$, or the input feature for the neural network, can be defined in several ways depending on the available sensors in a certain application. In this work, we consider two kinds of observations. In a simulated environment, we use the heightfield of the liquid surface, which can be easily computed from an exact geometrical representation of $\E{L}$, such as a set of particles. However, the dynamics of the liquid body are captured largely by the velocity field instead of its shape. To recover this information, we maintain a short memory of the heightfield over the past 4 frames as input to the neural network so that velocity information can be recovered by finite difference. This is illustrated in \prettyref{Fig:feature} (a). Longer-term memory can also be recovered using, e.g., recurrent neural network or LSTM. However, according to the definition of velocity as the derivative of positions, only the most recent memory is needed and these structures are unnecessary. However, in a real-life robotic system, acquiring the heightfield may be very difficult. As a result, we also experimented with a simplified feature, which is the height of liquid surface only at the lip of source container $\E{S}$, as illustrated in \prettyref{Fig:feature} (b). A common drawback of the heightfield feature is that we can only represent laminar flow without internal air bubbles in the liquid. However, we can carefully design our training dataset to contain only laminar flow (see \prettyref{sec:results} for more details).

\subsection{TRAINING DATA GENERATION}
The dataset used to train the neural network is difficult to acquire. This dataset should contain only successful pouring trajectories. Moreover, the neural network should be expressive enough to regenerate these trajectories. Previous works \cite{AISTATS2011_RossGB11,2014-cgps} address this problem for some applications by modifying the dataset during training. \cite{AISTATS2011_RossGB11} assumes there is an expert who can provide additional training samples on request for error recovery. However, our approach does not assume the presence of any such expert. A human demonstrator may serve as a good expert, but digitizing or capturing the liquid shape trajectory can be very challenging. On the other hand, \cite{2014-cgps} assumes that the governing \prettyref{eq:NS} is differentiable with respect to $u$. This assumption does not hold due to the non-smooth free-surface changes. Moreover, the computational complexity of \cite{2014-cgps} makes it infeasible for the high-dimensional configuration space of a liquid body.
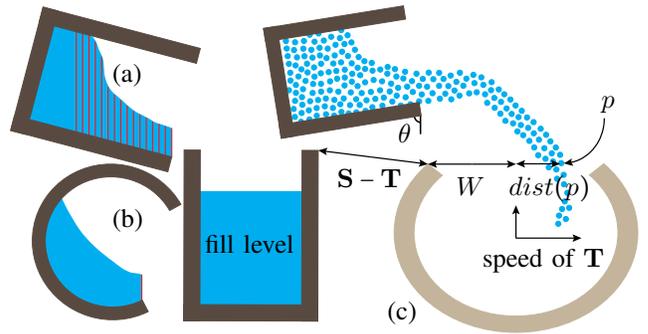
\begin{figure}
\begin{center}
\includegraphics[trim=0mm 0mm 0mm 0mm, clip, width=.98\linewidth]{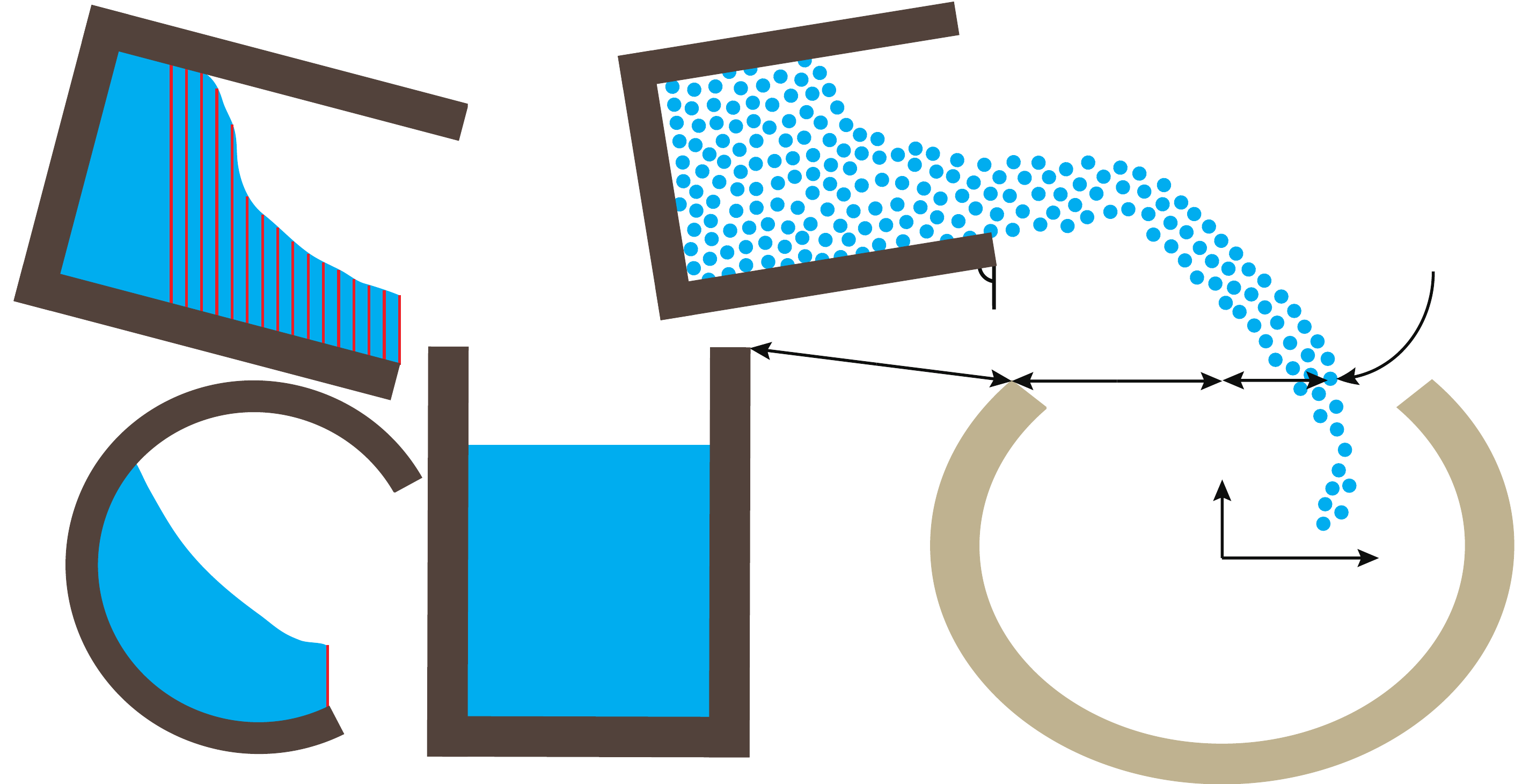}
\put (-200,95  ) {(a)}
\put (-200,40  ) {(b)}
\put ( -96,5   ) {(c)}
\put ( -92,72  ) {$\theta$}
\put ( -70,52  ) {$W$}
\put ( -50,52  ) {$dist(p)$}
\put ( -15,85  ) {$p$}
\put (-165,30  ) {fill level}
\put (-115,55  ) {$\E{S}-\E{T}$}
\put ( -60,25  ) {speed of $\E{T}$}
\end{center}
\vspace{-5px}
\caption{\label{Fig:feature} \changed{The two kinds of features we use. (a): The heightfield of the liquid free-surface (red). This feature is used as groundtruth in a simulated environment. (b): The height of liquid at the lip of source container $\E{S}$ (red). This feature makes it easier to apply our method on a real-life robotic system and on different container shapes. (c): The variables that define one instance of our pouring problem: relative position $\E{S}-\E{T}$, speed of $\E{T}$, and fill level. We also illustrate the variables that define our reward function $R$: The container opening width $W$ and the distance from a particle to a center point $dist(p)$.}}
\vspace{-15px}
\end{figure}

Our solution is to use stochastic optimization to automatically search for successful pouring trajectories similar to \cite{kuriyama2008trajectory} in 2D workspaces. We introduce several kinds of variations so that the learned neural network can be generalized to different problems. As illustrated in \prettyref{Fig:feature} (a), each pouring problem can be specified by three variables:
\begin{small}
\begin{itemize}
    \item The relative position $\E{S}-\E{T}$ in range $[-3,0]\times[-3,3](m)$.
    \item The constant moving speed of $\E{T}$ in range $[-5,5]^2(m/s)$.
    \item The liquid fill level of $\E{S}$ in range $[0.3,0.8]$ where $1$ means fully filled.
\end{itemize}
\end{small}
In order to quickly find a large number of successful pouring trajectories, we design a problem specific search space and reward function. Our liquid simulator requires very small timestep size $(\Delta t<0.01s)$ to ensure accuracy, leading to a large number of timesteps. We first limit the number of variables by using spline interpolation with $6$ control points for source container trajectory: $\E{S}(iK\Delta t/5)$ where $0\leq i\leq 5$. In 2D workspaces, each rigid configuration of $\E{S}$ consists of 2-dimensional translation and orientation $(\E{x}_\E{S},\E{y}_\E{S},\theta_{\E{S}})$, leading to a $15$ dimensional search space (the initial control point is fixed). However, we found that this is still a too large search space in practice because the optimizer can frequently generate zig-zag trajectories contrary to our intuitive observation of human pouring behaviour. Therefore, we further restricting the search space by observing that source container $\E{S}$ is always moving closer to $\E{T}$ and its turning angle is always increasing. This gives the following relationship:
\begin{tiny}
\begin{eqnarray*}
(\alpha_i\triangleq\frac{|\E{x}_\E{S}-\E{x}_\E{T}|_i}{|\E{x}_\E{S}-\E{x}_\E{T}|_{i-1}},
\beta_i\triangleq\frac{|\E{y}_\E{S}-\E{y}_\E{T}|_i}{|\E{y}_\E{S}-\E{y}_\E{T}|_{i-1}},
\gamma_i\triangleq\frac{|\theta_{max}-\theta|_i}{|\theta_{max}-\theta|_{i-1}}).
\end{eqnarray*}
\end{tiny}
We propose to search in the transformed coordinates $(\alpha_i,\beta_i,\gamma_i)\in(0,1]^3$ using the CMA-ES algorithm \cite{hansen2016cma}. Although this is still a $15$ dimensional seach space, much fewer random samples are needed in each iteration with such transformation. Finally, we use the following reward function for CMA-ES optimization that encourages particle to pass through the center of target container opening and penalize spillage:
\begin{small}
\begin{eqnarray*}
R=\sum_{p}R_{p}\quad R_{p}=
\begin{cases}
   \frac{W-dist(p)}{W},  &  \text{if } dist(p)<W\\
   -100,                 &  \text{otherwise}
\end{cases}
\end{eqnarray*}
\end{small}
where $p$ loops over all particles, $dist(p)$ is its distance to center of opening of $\E{T}$, and $W$ is the half width of opening, as illustrated in \prettyref{Fig:feature} (b). 

After we find the optimal $\E{S}(t)$ in our search space, we extract groundtruth observation $O(\E{L}(t))$ and label $(\E{S}'(t),p,v_x,v_y,\rho)$ for each spline interpolated timestep. To extract the quadratic curve parameters, we use the same greedy quadratic curve fitting method as \cite{pan2016robot}. Finally, our neural-network outputs the transformed coordinates $(\alpha_i,\beta_i,\gamma_i)$ instead of $(\E{x}_\E{S},\E{y}_\E{S},\theta_{\E{S}})$ that fits in the range of sigmoid activation function.
\end{changedBlk}

\begin{changedBlk}
\section{ANALYSIS AND RESULTS}\label{sec:results}
In this section, we evaluate the online and offline phases in \prettyref{Fig:teaser} from different aspects.

\TE{Quality of Dataset:} All the trajectories in our dataset reside in 2D workspaces. The \CC{} of generating these trajectories is: $\mathcal{O}(I\times R\times K\times P)$, where $I$ is the number of iterations needed in each CMA-ES optimization, $R$ is the number of random samples used in each CMA-ES iteration, $K$ is the number of timesteps in each fluid simulation, and finally $P$ is the number of trajectories $\E{S}(t)$ we want to generate. In all the optimizations, we set $I=200$, $R=30$, $K=500$, and $\Delta t=0.01(sec)$. In other words, each trajectory lasts for $K\Delta t=5(sec)$. The \CC{} of dataset generation is also proportional to the overhead of performing liquid simulation, i.e., evaluating function $f$, which in turn is proportional to the number of particles. In 2D workspaces, we use approximately $10^5$ particles and each evaluation of $f$ takes $2.5(sec)$; while the number of particles in 3D workspaces is $10^7$ and each evaluation takes $345(sec)$. Therefore, generating a 3D dataset is orders of magnitude more expensive than a 2D dataset.
\begin{figure}[t]
\begin{center}
\includegraphics[trim=0mm 0mm 0mm 0mm, clip, width=.95\linewidth]{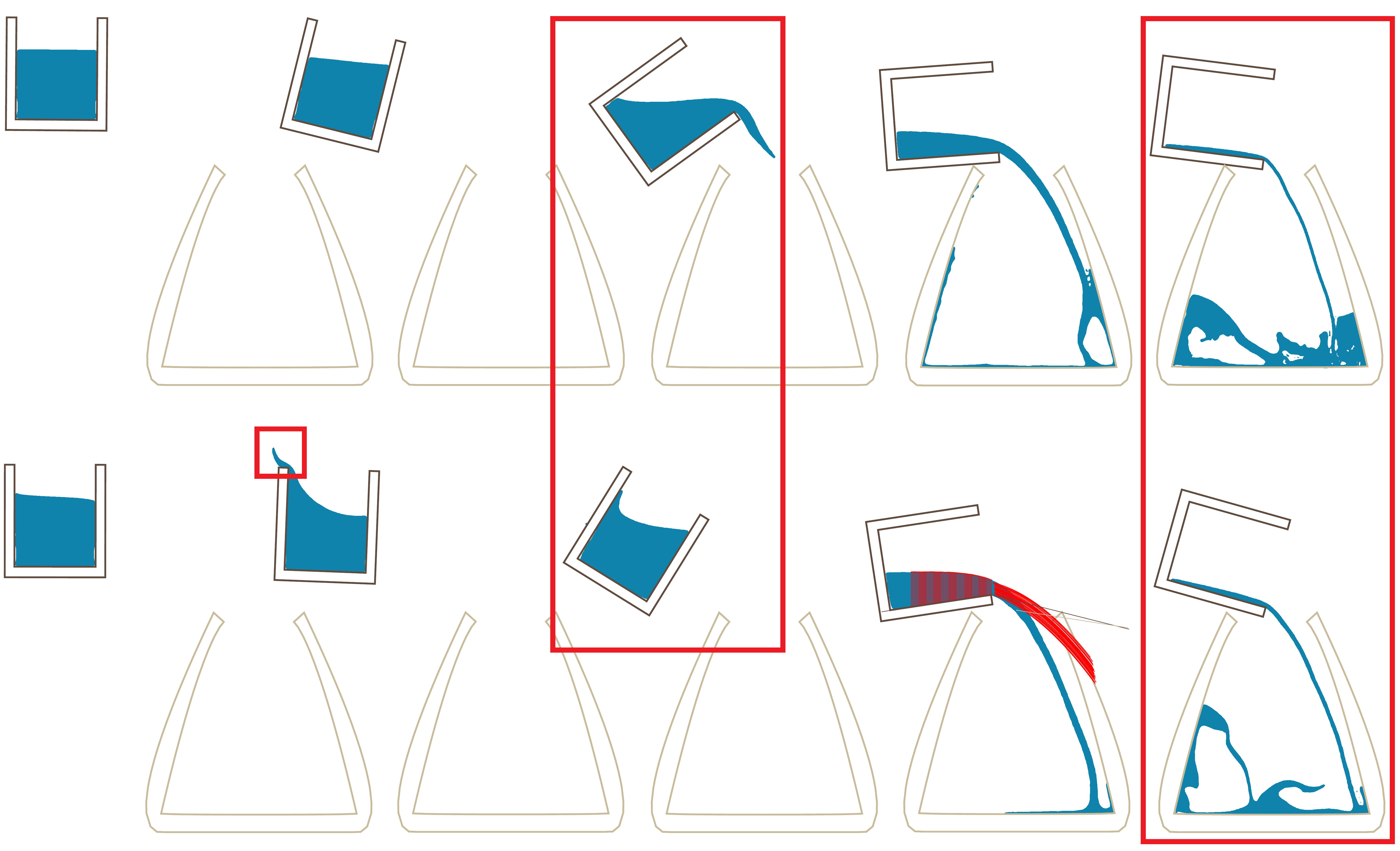}
\put (-230,105) {(a)}
\put (-230,30 ) {(b)}
\put (-20 ,65 ) {(c)}
\put (-180,68 ) {(d)}
\put (-120,65 ) {(e)}
\put (-60 ,50 ) {(f)}
\end{center}
\caption{\label{Fig:DSIllus} We illustrate an exemplary trajectory of \TDS{} (a) and \TSDS{} (b). On convergence of CMA-ES optimization, the liquid flow is well centered around the opening of $\E{T}$ (c). \TSDS{} encourages spillage at an early stage of pouring (d), so that $\E{S}$ must move and turn slowly (e). For each timestep in each trajectory, we extract the ground truth water height field and \LOP{} as training dataset (f).}
\vspace{-10px}
\end{figure}

\begin{wrapfigure}{r}{0.5\linewidth}
\vspace{-5px}
\begin{center}
\includegraphics[trim=0mm 0mm 0mm 0mm, clip, width=1.0\linewidth]{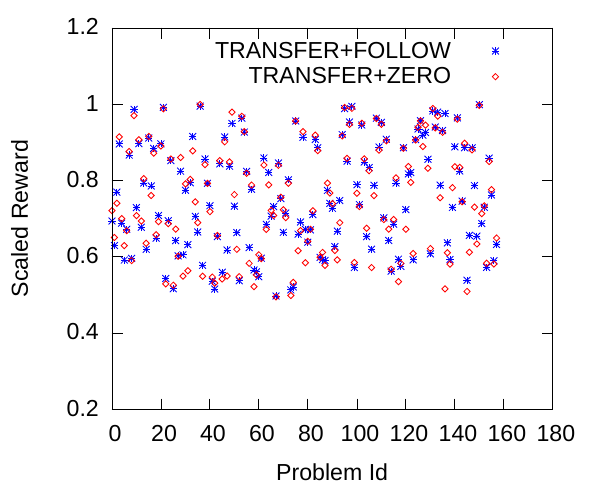}
\end{center}
\vspace{-5px}
\caption{\label{Fig:DSReward} The scaled reward function $R/\#Particle$ for a set of transfer problems in \TDS{} (blue) and \TSDS{} (red). These values are all positive, meaning that very little spillage happens and particles are well centered around $\E{O}_\E{T}$.}
\vspace{-5px}
\end{wrapfigure}
We generated two datasets named \TDS{} and \TSDS{} each containing $P=1000$ successful pouring trajectories. Specifically, we first sample the fill level with interval $0.12$ and then select $200$ relative positions $\E{S}-\E{T}$ and moving speed of $\E{T}$ for each fill level using uniform random sampling in the given range. The generation of \TDS{} and \TSDS{} involves $6\times10^9$ evaluations of $f$ altogether. \TDS{} and \TSDS{} differ in the initial liquid configuration. In \TDS{}, the initial liquid velocity follows that of $\E{S}$, which is typical if the liquid has moved with $\E{S}$ for a distance and reached equilibrium. However, in \TSDS{}, the initial liquid velocity is zero, which is typical if we start transfer from a stationary scenario. Problems in \TSDS{} are considered harder than those in \TDS{}, as moving $\E{S}$ too quickly will lead to spillage and thus negative reward. These two datasets can be downloaded at \href{http://gamma.cs.unc.edu/RLFluid}{our project page} and are illustrated in \prettyref{Fig:DSIllus}. 

\prettyref{Fig:DSReward} shows the distribution of scaled reward function $R/\#Particle$ in \TDS{} and \TSDS{}. \prettyref{Fig:DSConvergence} shows the CMA-ES convergence history. These figures show that our stochastic optimization algorithm can efficiently find successful transfer trajectories. Furthermore, we need to verify that the generated trajectories always transfer liquid using laminar flow, instead of turbulent flow, so that a heightfield feature can represent the shape of $\E{L}$. We note that the velocity field of a laminar flow should have no internal vortex. Therefore, we compute the vortical velocity component and plot its strength relative to the original velocity field in \prettyref{Fig:VelComp}. This figure shows clearly that the trajectories in both \TDS{} and \TSDS{} use only laminar flow.

\begin{figure}
\begin{center}
\includegraphics[trim=0mm 0mm 0mm 0mm, clip, width=.49\linewidth]{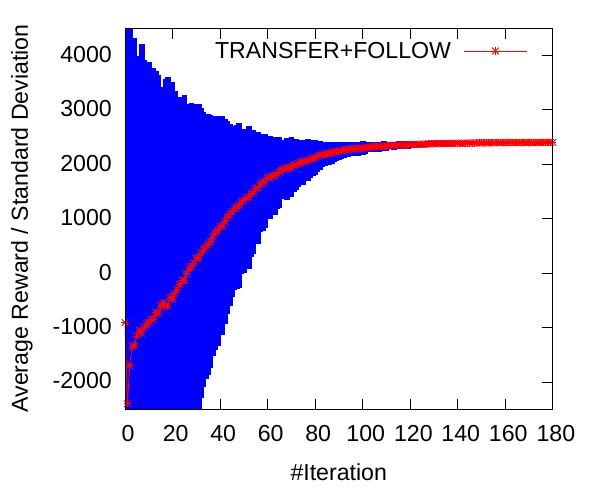}
\includegraphics[trim=0mm 0mm 0mm 0mm, clip, width=.49\linewidth]{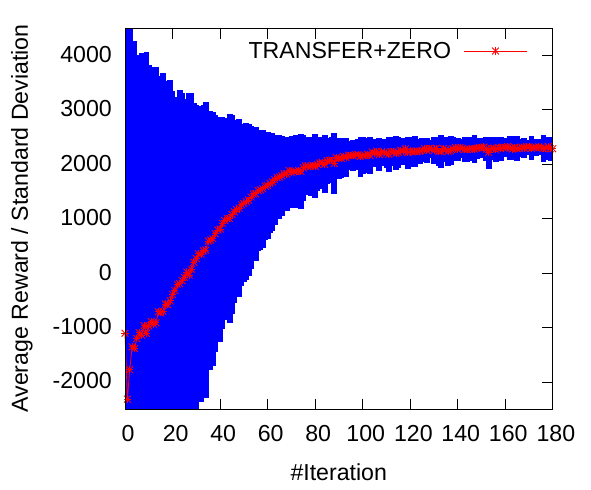}
\put (-150,30 ) {(a)}
\put (-25 ,30 ) {(b)}
\end{center}
\vspace{-5px}
\caption{\label{Fig:DSConvergence} \changed{The average convergence history of CMA-ES algorithm over $1000$ problems. This algorithm converges equally well for \TDS{} (a) and \TSDS{} (b).}}
\vspace{-10px}
\end{figure}

\begin{figure}
\begin{center}
\includegraphics[trim=0mm 0mm 0mm 0mm, clip, width=.98\linewidth]{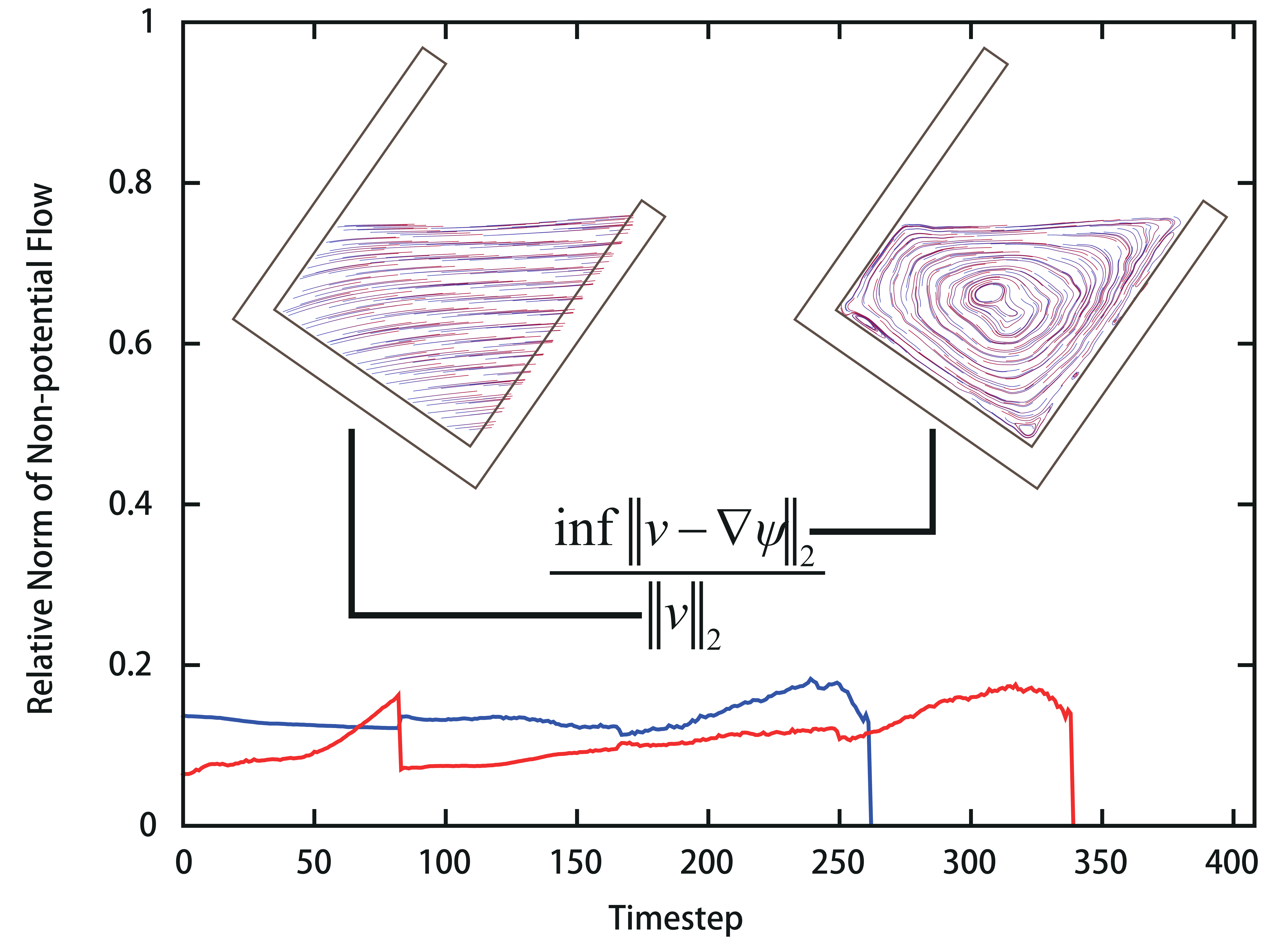}
\end{center}
\vspace{-5px}
\caption{\label{Fig:VelComp} \changed{We picked 100 random trajectories in \TDS{} (blue) and \TSDS{} (red) dataset, and visualized the temporal change of the relative strength of vortical velocity component (the numerator $\E{inf}\|v-\nabla\psi\|_2$). Since this value is always less than $0.2$, the flow is very close to a potential flow ($\E{inf}\|v-\nabla\psi\|_2=0$). Since problems in the \TDS{} dataset are easier, the pouring is usually completed faster, which is consistent with the early termination of the blue curve.}}
\vspace{-10px}
\end{figure}
\end{changedBlk}

\begin{changedBlk}
\TE{Accuracy of Neural Network:} Our neural network serves two purposes, that are verified with separate experiments. 

We first verify the accuracy of the \LOP{} predictor by testing it on $100$ randomly selected trajectories in \TDS{}/\TSDS{}, i.e. testing using the training dataset. Next, we test it on $100$ new trajectories $\mathcal{P}$ held out from training data. \prettyref{Fig:LOPPred} (a) plots the averaged accuracy against the turning angle $\alpha$. This plot reveals that our predictor has low accuracy at a small turning angle (up to $0.46\%$ at $40^\circ\pm10^\circ$). The average relative accuracy throughout pouring process is $15\%$ using the heightfield feature,\prettyref{Fig:feature} (a), and $24\%$ using the height at lip feature, \prettyref{Fig:feature} (b). At small turning angle, using height at lip feature increase the error by $36\%$ for \TDS{} dataset, while the performance is almost the same for \TSDS{} dataset. At larger turning angle, using height at lip feature only increases the error by $10\%$ at most.
\begin{figure}
\begin{center}
\includegraphics[trim=0mm 0mm 0mm 0mm, clip, width=.48\linewidth]{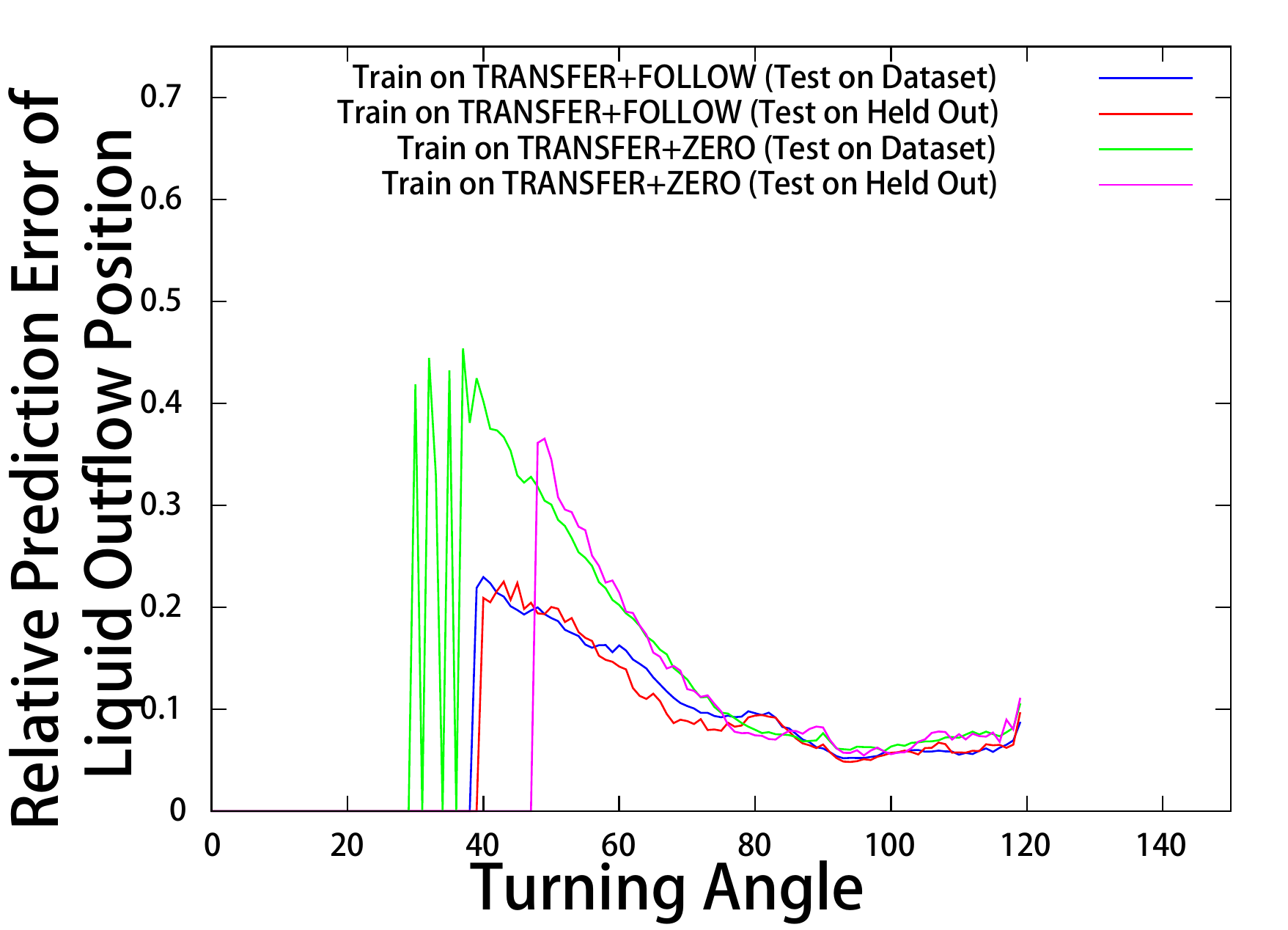}
\includegraphics[trim=0mm 0mm 0mm 0mm, clip, width=.48\linewidth]{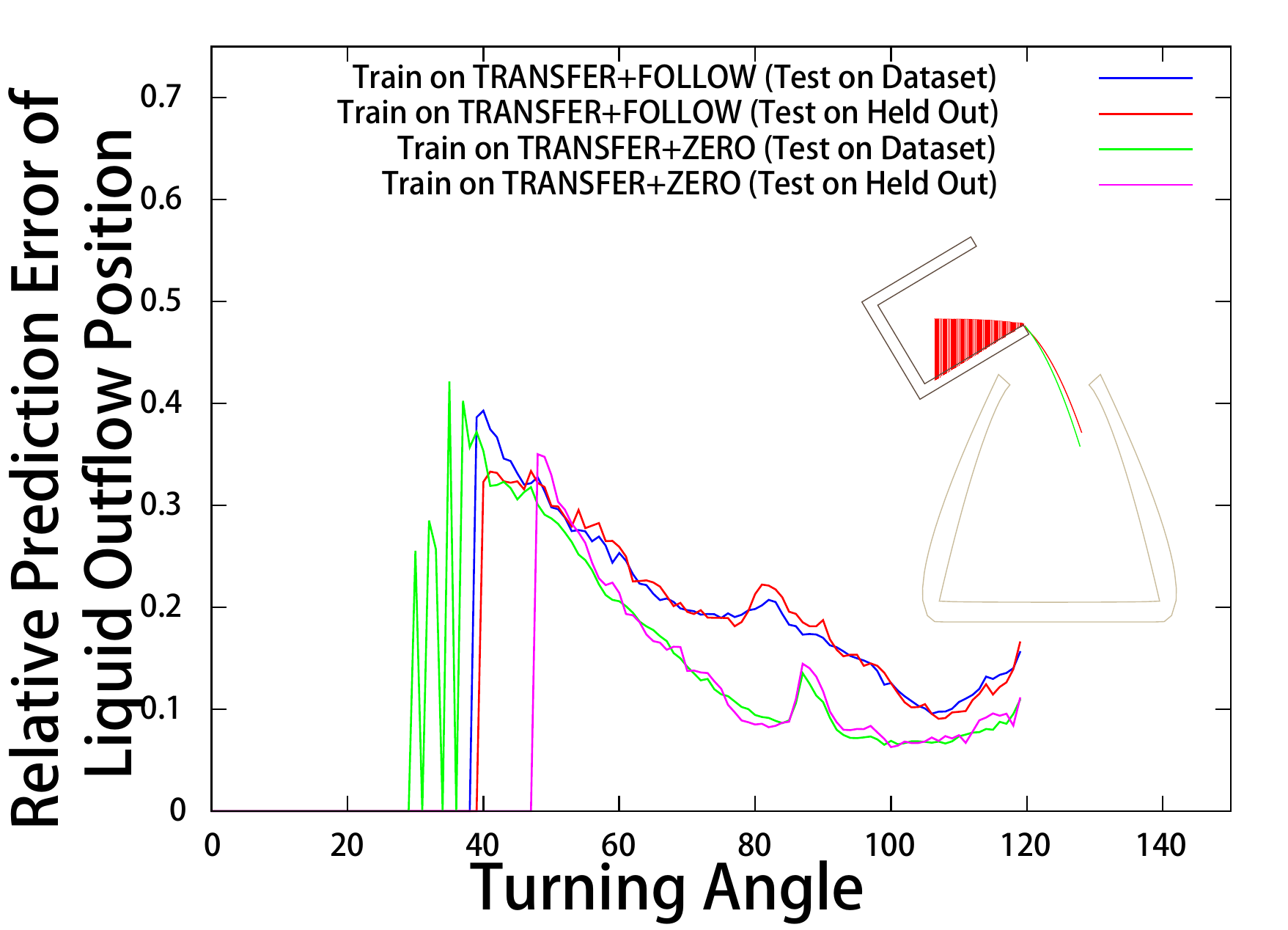}
\put (-185,45 ) {(a)}
\put (-60 ,45 ) {(b)}
\end{center}
\vspace{-5px}
\caption{\label{Fig:LOPPred} \changed{Relative error of predicted \LOP{} parameters ($\|v_x,v_y,p\|^2$) using heightfield feature (a) and height at lip feature (b). The error is plotted against turning angle of $\E{S}$.}}
\vspace{-10px}
\end{figure}

We further validate the accuracy of the \MTP{} predictor. \prettyref{Fig:MTPPred} (a) illustrates the relative accuracy of $\E{S}(t)$ compared with the groundtruth. The error is higher than that of the \LOP{} predictor. One of the main purposes of our online feedback motion planner is to compensate for this error. In addition, we conduct another experiment to see if our predictor has the ability to avoid spillage. Since we know that \TDS{} does not model spillage but \TSDS{} can model spillage, we plot the predicted temporal change of $\alpha$ using \TDS{} and \TSDS{} as the training dataset. Our predictor trained using the \TSDS{} dataset prefers a lower $\dot\alpha$ at small $\alpha$, which implies that our network can learn the ability to avoid spillage.
\begin{figure}
\begin{center}
\includegraphics[trim=0mm 0mm 0mm 0mm, clip, width=.51\linewidth]{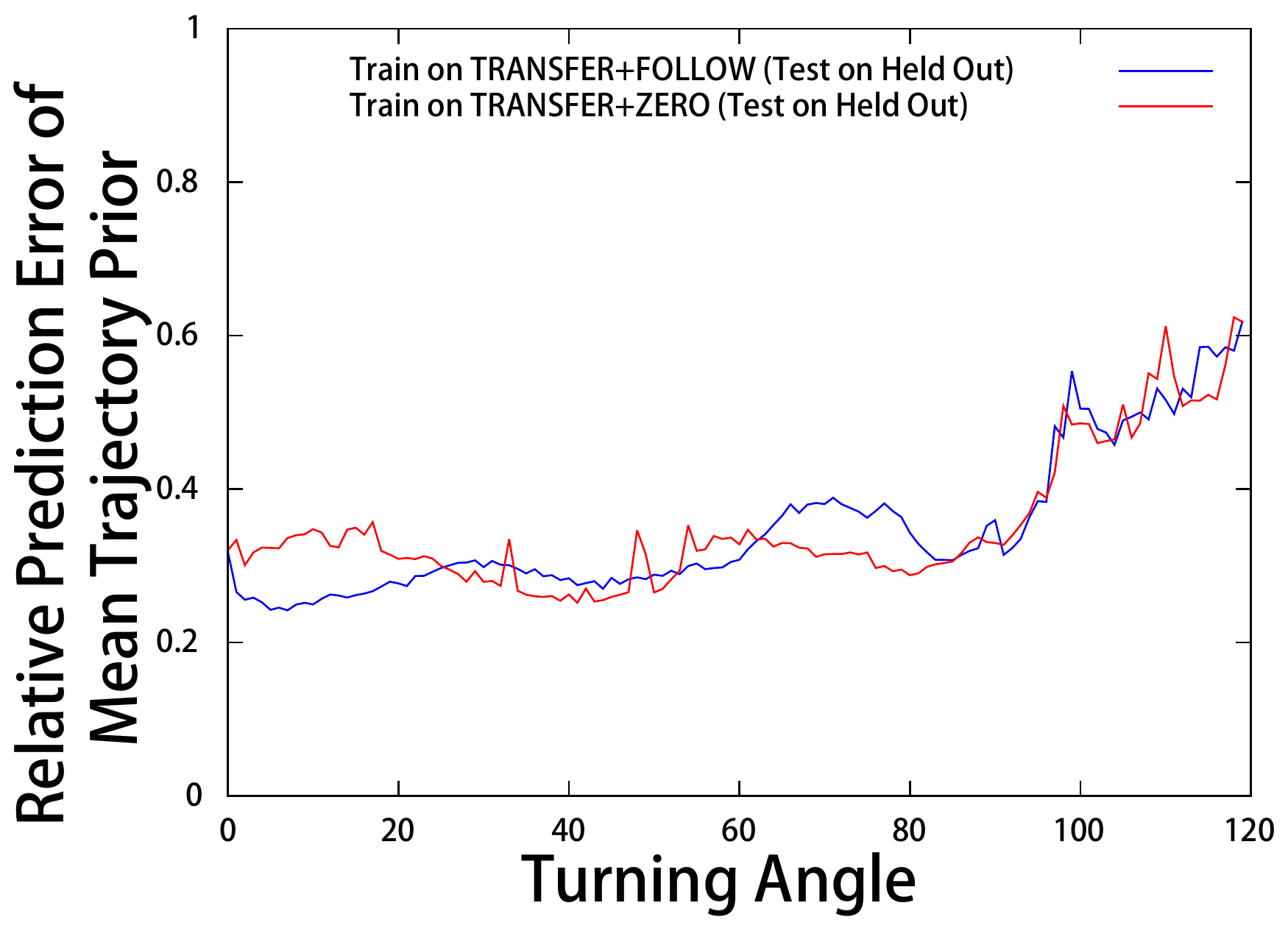}
\includegraphics[trim=0mm 0mm 0mm 0mm, clip, width=.47\linewidth]{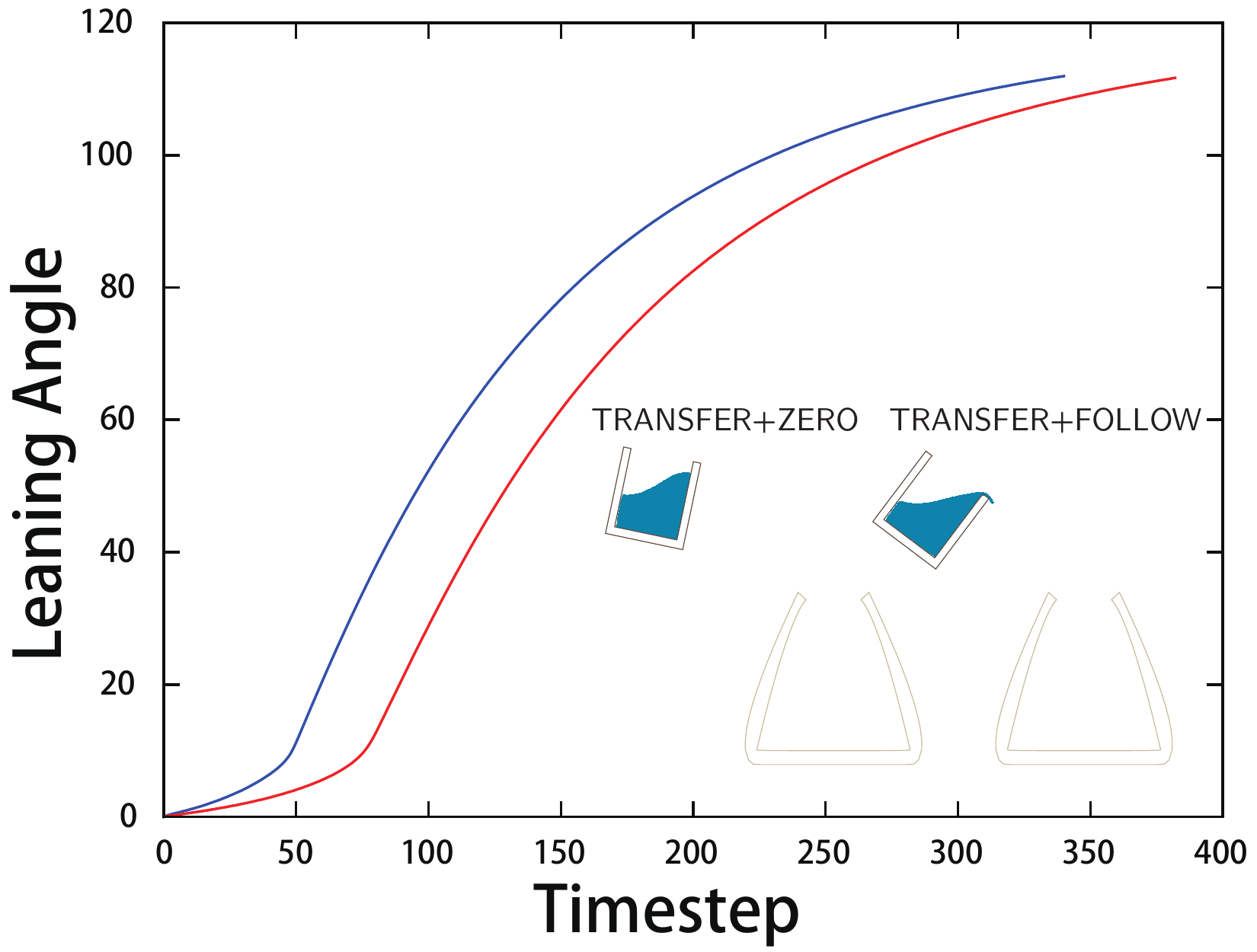}
\put (-170,20 ) {(a)}
\put (-70 ,20 ) {(b)}
\end{center}
\vspace{-5px}
\caption{\label{Fig:MTPPred} \changed{(a): The relative error of predicted $\E{S}(t)$ plotted against turning angle. (b): The predicted turning angle of $\E{S}$ against timestep index: the \MTP{} predictor trained using \TSDS{} dataset (red) does learn to turn $\E{S}$ slowly to avoid spillage, compared with the predictor trained using \TDS{} dataset (blue). We also show such difference using one exemplary frame of a testing problem.}}
\vspace{-15px}
\end{figure}
\end{changedBlk}

\begin{figure*}[t]
\begin{center}
\includegraphics[trim=0mm 0mm 0mm 0mm, clip, width=.98\linewidth]{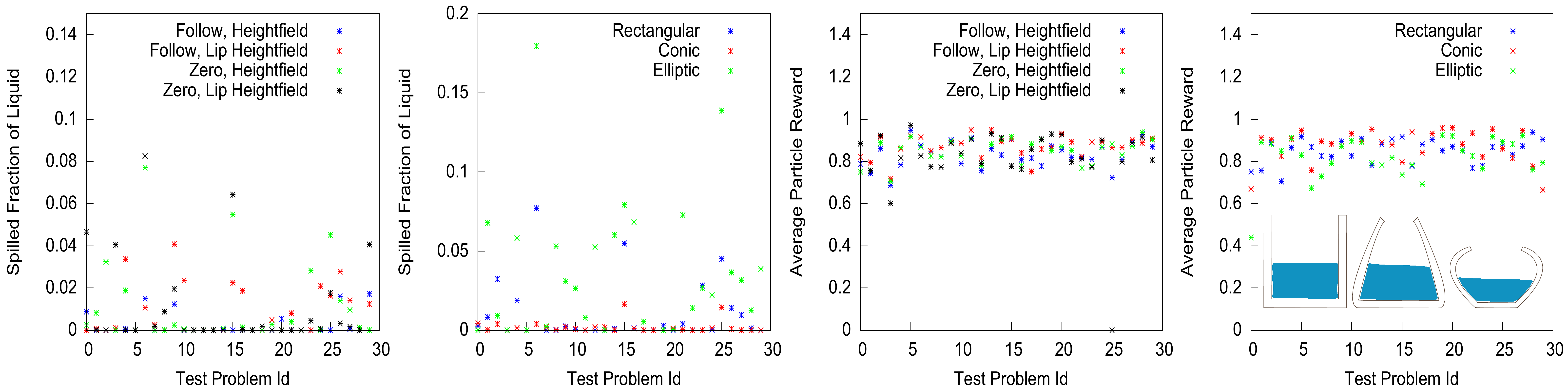}
\end{center}
\vspace{-10px}
\caption{\label{Fig:FDPerf} \changed{Performance of feedback planner. (a): The fraction of spilled particle using rectangular container $\E{S}$, two kinds of datasets and features. (b): The fraction of spilled particle using height at lip feature and three different container shapes. (c): Average reward of particles that fall into $\E{T}$ in experiment (a). (d): Average reward of particles that fall into $\E{T}$ in experiment (b).}}
\vspace{-5px}
\end{figure*}
\begin{changedBlk}
\TE{Performance of Feedback Planner:} In the online phase, we solve \prettyref{eq:Opt} using a horizon length of $1.25(sec)$, with $K=25$ and $\Delta t=0.05(sec)$. This requires querying the neural network 25 times and performing an optimization using the LBFGS optimizer, which is very efficient due to the small size of our neural network and the optimization formulate. We tested our feedback controller on $30$ new problems. For each testing problem, we experimented with three different container shapes as illustrated in \prettyref{Fig:FDPerf} (d). We use the same set of $30$ experimental problems which are not covered in the datasets. If the training is accomplished using \TSDS{} dataset then we set the initial velocity to be zero and vice versa. And for experiments using new container shapes, we use \TSDS{} dataset.

The performance of planner is summarized in \prettyref{Fig:FDPerf} where we plot the spilled fraction of liquid and average reward. Note the the average is taken over particles that fall into $\E{T}$, i.e., excluding the spilled particles, which is different from \prettyref{Fig:DSReward} where the average includes negative values (spilled particles). In this way, we can analyze spillage avoidance and the accuracy of liquid flow, respectively. From \prettyref{Fig:FDPerf} (a), we can see that the fraction of spilled particles is less than $5\%$ in $28$ of $30$ problems for both features in \prettyref{Fig:feature} and using lip height feature does increase the spillage by $4\%$ at most and $0.32\%$ on average. From \prettyref{Fig:FDPerf} (b), we find that, although our datasets use only rectangular container, the spillage fraction is also very small if we test on conic container, with spillage fraction over $5\%$ for only $2$ problems. However, some container, such as the conic container, encourages spillage and we have seen over $5\%$ spillage in $10$ problems. From \prettyref{Fig:FDPerf} (c), we can see that the liquid flow are generally well centered around the center of opening of $\E{T}$ with a mean reward of $R_p=0.82$. If we generalize to new container shapes, the mean reward is still over $R_p=0.8$ but the variance is larger especially for the conic container, as illustrated in \prettyref{Fig:FDPerf} (d). 
\end{changedBlk}

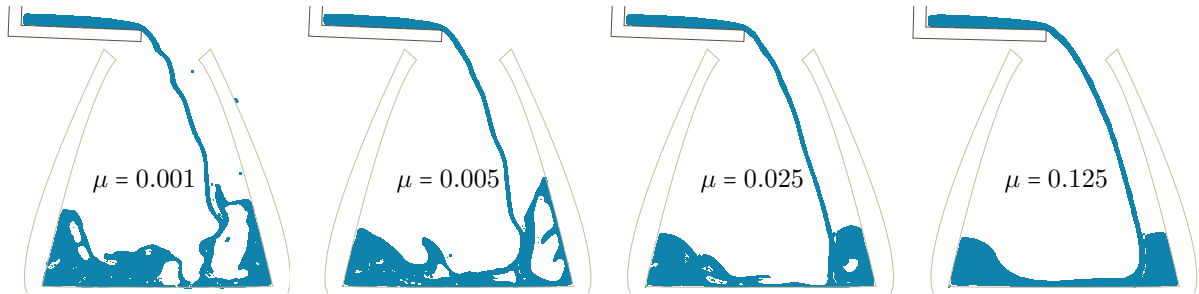
\begin{figure*}[t]
\begin{center}
\includegraphics[trim=0mm 0mm 0mm 0mm, clip, width=.9\linewidth]{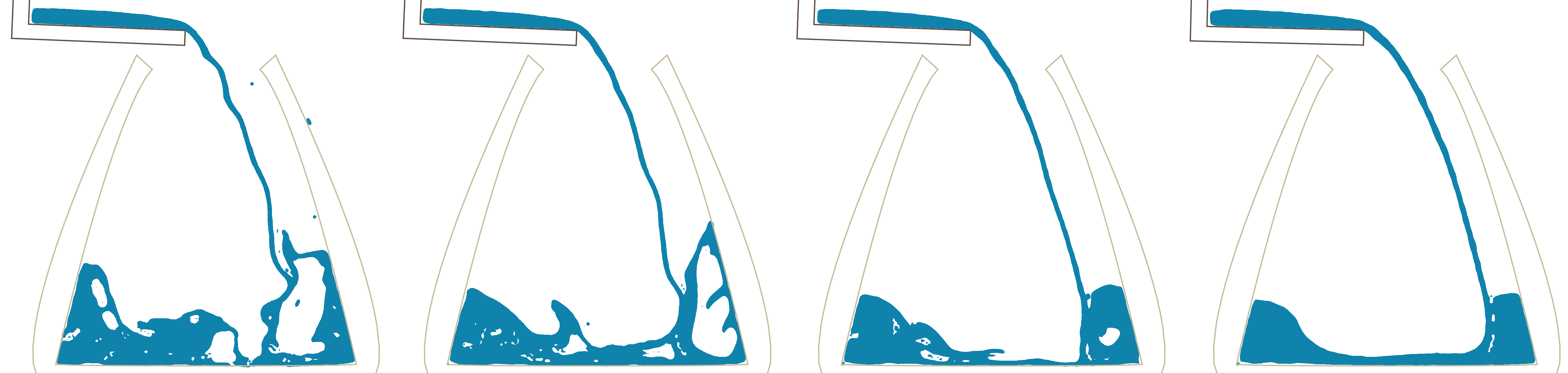}
\put (-420,40 ) {$\mu=0.001$}
\put (-305,40 ) {$\mu=0.005$}
\put (-190,40 ) {$\mu=0.025$}
\put (-75 ,40 ) {$\mu=0.125$}
\end{center}
\caption{\label{Fig:HighVisc} We tested our feedback controller on new fluid materials with higher dynamic viscosity ($kg/(ms)$). In this case, we achieved even higher online reward.}
\vspace{-15px}
\end{figure*}

\begin{changedBlk}
\TE{Generalization to New Problems:} Our motion planner is only experimented on liquid body of fixed material parameter, 2D workspace, and two rigid bodies. Some of these limitations can be roughly relaxed. 

First, although our dataset considers only liquid material with low viscosity, applying it to pour more viscous liquids is possible. As illustrated in \prettyref{Fig:HighVisc}, we experimented the planner on these liquids. As we increase the viscosity, the fraction of spillage decreases and the average reward increases accordingly. This is because viscosity suppresses turbulence flow and makes our quadratic curve assumption more accurate. 

Moreover, we can also relax the assumption of 2D workspaces and apply the method to 3D workspaces. As illustrated in \prettyref{Fig:3DCase} (b), this is done by applying the method to only the symmetric cross section. A drawback of this treatment is that fluid may spill from other directions than the 2D plane and our method cannot avoid such spillage at all. 

Finally, a major benefit of the optimization-based formulation, \prettyref{eq:Opt}, is that we can naturally take other dynamic obstacles into consideration without retraining or regenerating the dataset, as illustrated in \prettyref{Fig:3DCase} (a). However, current obstacle avoidance term only take rigid bodies into consideration, and collisions between liquid and rigid bodies are not considered. As a result, when the obstacle blocks the way of pouring, more spillage happens.

\begin{figure}
\begin{center}
\includegraphics[trim=0mm 0mm 0mm 0mm, clip, width=.9\linewidth]{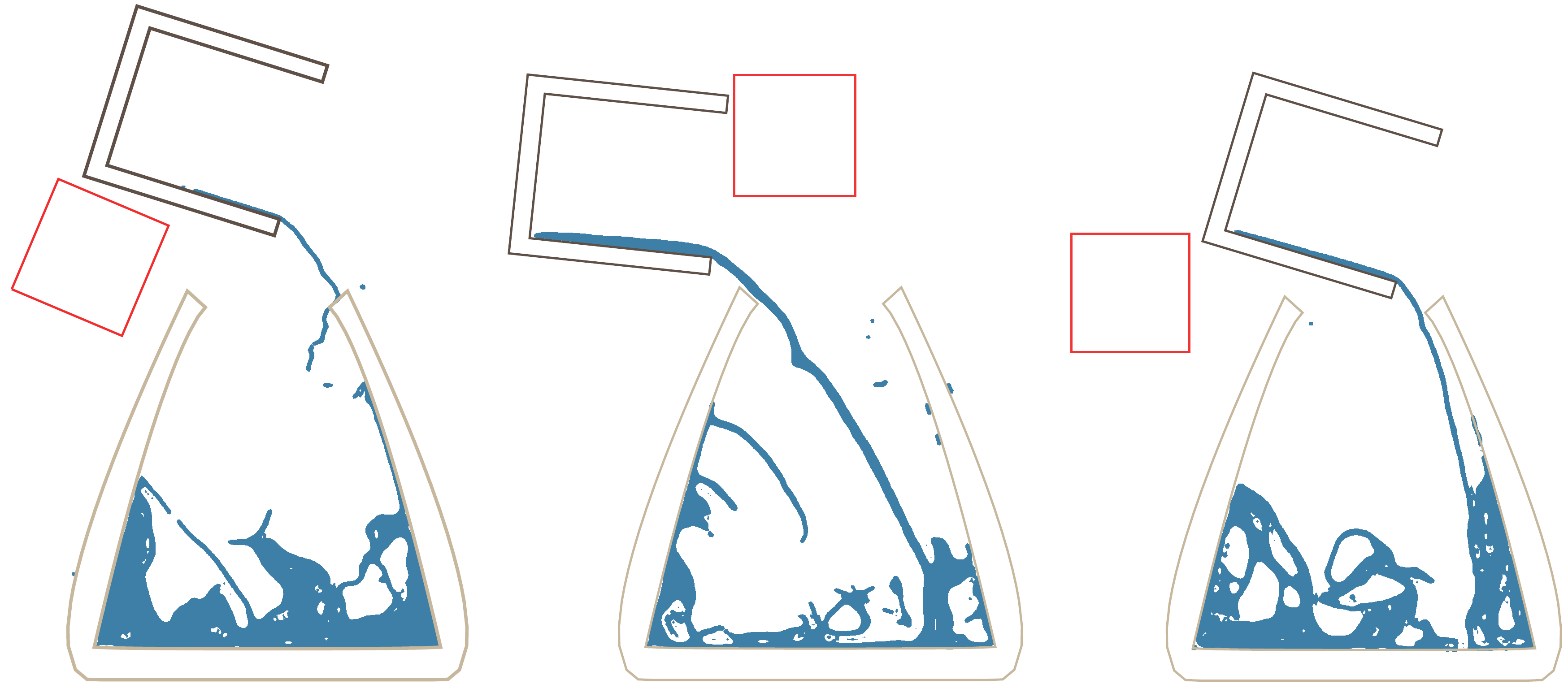}
\includegraphics[trim=0mm 0mm 0mm 0mm, clip, width=.95\linewidth]{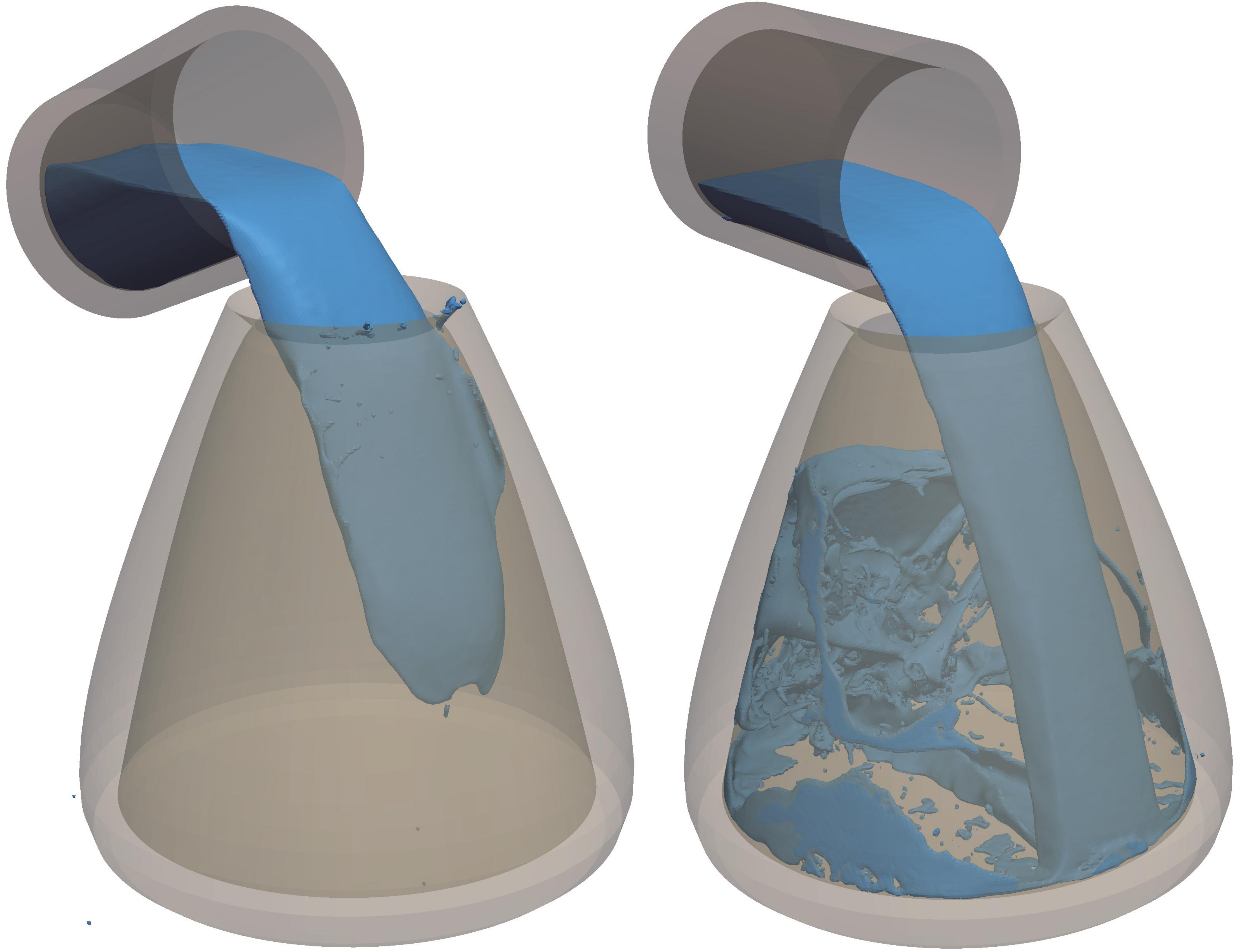}
\put (-20 ,180) {(a)}
\put (-20 ,-5 ) {(b)}
\end{center}
\caption{\label{Fig:3DCase} (a): Examples with dynamic obstacles. (b): A frame of our controller applied to a 3D workspace. The controller only sees the 2D cross section.}
\vspace{-10px}
\end{figure}
\end{changedBlk}

\section{LIMITATIONS AND CONCLUSIONS}\label{sec:conc}
\begin{changedBlk}
In conclusion, we present a feedback motion planner for liquid transfer problems. Our formulation uses an optimization-based receding horizon planner, which is guided by a machine learning model that provides clues for global movements (\MTP{}) and local adjustments (\LOP{}). Our experiments show that the planning framework can achieve promising online performance and the machine learning model gains important skills in pouring such as spilling avoidance and liquid position prediction. However, current system also introduces a series of limitations and possible future works discussed below.

\TE{Better Performance:} Although the spillage fraction in our dataset is almost zero, the spillage fraction can be as high as $5\%$ in our $30$ testing problems. Several approaches can be used to further improve the performance. One possible method is to combine new features to improve the accuracy of machine learning model, such as liquid related feature \cite{pan2016simple} and geometrical feature \cite{brandi2014generalizing}. 

Another method is to use imitation learning \cite{AISTATS2011_RossGB11} instead of supervised learning, so that the dataset and neural network becomes more compatible. However, a drawback of this method is that the reusability of the dataset is compromised. Due to the high computational cost of liquid simulation, relying on a reusable dataset such as \TDS{} and \TSDS{} is an advantage.

\TE{Generalization to Other Materials:} It is inherently difficult to generalize our method to discrete material types, such as a bunch of rigid bodies or granular materials. This is because the quadratic outflow curve assumption is valid only if the material can be modelled as a continuum. In this sense, our method is disadvantageous compared with \cite{yamaguchi2014learning,yamaguchi2016neural}. For granular materials, identifying a low-dimensional parameters for learning is a challenging future work.

\TE{More Complex Manipulation Tasks:} In this work, we only consider the problem of pouring the entire liquid body from $\E{S}$ to $\E{T}$. However, other requirements might arise, e.g., if only part of liquid are needed in $\E{T}$ due to the small volume of $\E{T}$. Such generalization can be considered as future work by slowly decreasing the turning angle when the weight of $\E{S}$ is smaller than some threshold. Another situation is that we need to shake the source container to force jelly liquids out of $\E{S}$ with narrow opening, for which \cite{yamaguchi2014learning,yamaguchi2016neural} is a better choice than trajectory optimization. Finally, a robotic arm might need to transverse through environments with complex dynamic obstacles in order to reach $\E{T}$. Instead of simply using a collision cost in \prettyref{eq:Opt}, which leads to more spillage as shown in \prettyref{Fig:3DCase} (a), a better method is to discontinue the pouring until the obstacle moves away. Such multi-stage pouring is considered as future work.

\TE{Application to Real Robotic System:} Two problems need to be addressed for extension to real robotic systems. First, we must be able to efficiently acquire liquid-related features. Although we have shown that reasonable performance can be achieved using height at lip feature, it is still difficult to acquire due to the transparent appearance of liquid bodies. A possible workaround is to remember the geometry of $\E{S}$ and compute the height at lip using the geometrical relationship in \cite{pan2016simple}. 

Another challenge is to extend the entire planning pipeline to 3D workspaces, for which dataset generation can be extremely costly. Indeed, even a single 3D liquid simulation over $5(sec)$ takes $6(hr)$ on our desktop machine, making millions of liquid simulations impractical. We are currently looking at massive parallel machines for such computation.
\end{changedBlk}

\bibliographystyle{IEEEtran}
\nocite{*}
\bibliography{template}

\begin{thebibliography}{10}
\providecommand{\url}[1]{#1}
\csname url@rmstyle\endcsname
\providecommand{\newblock}{\relax}
\providecommand{\bibinfo}[2]{#2}
\providecommand\BIBentrySTDinterwordspacing{\spaceskip=0pt\relax}
\providecommand\BIBentryALTinterwordstretchfactor{4}
\providecommand\BIBentryALTinterwordspacing{\spaceskip=\fontdimen2\font plus
\BIBentryALTinterwordstretchfactor\fontdimen3\font minus
  \fontdimen4\font\relax}
\providecommand\BIBforeignlanguage[2]{{%
\expandafter\ifx\csname l@#1\endcsname\relax
\typeout{** WARNING: IEEEtran.bst: No hyphenation pattern has been}%
\typeout{** loaded for the language `#1'. Using the pattern for}%
\typeout{** the default language instead.}%
\else
\language=\csname l@#1\endcsname
\fi
#2}}

\bibitem{STOMP:2011}
M.~Kalakrishnan, S.~Chitta, E.~Theodorou, P.~Pastor, and S.~Schaal, ``{STOMP}:
  Stochastic trajectory optimization for motion planning,'' in
  \emph{Proceedings of IEEE International Conference on Robotics and
  Automation}, 2011, pp. 4569--4574.

\bibitem{schulman2014motion}
J.~Schulman, Y.~Duan, J.~Ho, A.~Lee, I.~Awwal, H.~Bradlow, J.~Pan, S.~Patil,
  K.~Goldberg, and P.~Abbeel, ``Motion planning with sequential convex
  optimization and convex collision checking,'' \emph{The International Journal
  of Robotics Research}, vol.~33, no.~9, pp. 1251--1270, 2014.

\bibitem{yamaguchi2016neural}
A.~Yamaguchi and C.~G. Atkeson, ``Neural networks and differential dynamic
  programming for reinforcement learning problems,'' in \emph{2016 IEEE
  International Conference on Robotics and Automation (ICRA)}.\hskip 1em plus
  0.5em minus 0.4em\relax IEEE, 2016, pp. 5434--5441.

\bibitem{yamaguchi2015differential}
------, ``Differential dynamic programming with temporally decomposed
  dynamics,'' in \emph{Humanoid Robots (Humanoids), 2015 IEEE-RAS 15th
  International Conference on}.\hskip 1em plus 0.5em minus 0.4em\relax IEEE,
  2015, pp. 696--703.

\bibitem{kuriyama2008trajectory}
Y.~Kuriyama, K.~Yano, and M.~Hamaguchi, ``Trajectory planning for meal assist
  robot considering spilling avoidance,'' in \emph{Control Applications, 2008.
  CCA 2008. IEEE International Conference on}.\hskip 1em plus 0.5em minus
  0.4em\relax IEEE, 2008, pp. 1220--1225.

\bibitem{pan2016robot}
Z.~Pan, C.~Park, and D.~Manocha, ``Robot motion planning for pouring liquids,''
  in \emph{Proceedings of International Conference on Automated Planning and
  Scheduling}, 2016.

\bibitem{kunze2011simulation}
L.~Kunze, M.~E. Dolha, E.~Guzman, and M.~Beetz, ``Simulation-based temporal
  projection of everyday robot object manipulation,'' in \emph{The 10th
  International Conference on Autonomous Agents and Multiagent Systems-Volume
  1}.\hskip 1em plus 0.5em minus 0.4em\relax International Foundation for
  Autonomous Agents and Multiagent Systems, 2011, pp. 107--114.

\bibitem{tzamtzi2008robustness}
M.~Tzamtzi and F.~Koumboulis, ``Robustness of a robot control scheme for liquid
  transfer,'' in \emph{Novel Algorithms and Techniques In Telecommunications,
  Automation and Industrial Electronics}.\hskip 1em plus 0.5em minus
  0.4em\relax Springer, 2008, pp. 156--161.

\bibitem{pan2016simple}
Z.~Pan and D.~Manocha, ``Motion planning for fluid manipulation using
  simplified dynamics,'' in \emph{IEEE/RSJ International Conference on
  Intelligent Robots and Systems}.\hskip 1em plus 0.5em minus 0.4em\relax IEEE,
  2016.

\bibitem{lavalle1998rapidly}
S.~M. Lavalle, ``Rapidly-exploring random trees: A new tool for path
  planning,'' 1998.

\bibitem{GRVO:2009}
D.~Wilkie, J.~P. van~den Berg, and D.~Manocha, ``Generalized velocity
  obstacles,'' in \emph{Proceedings of IEEE/RSJ International Conference on
  Intelligent Robots and Systems}, 2009, pp. 5573--5578.

\bibitem{stilman2007task}
M.~Stilman, ``Task constrained motion planning in robot joint space,'' in
  \emph{IEEE/RSJ International Conference on Intelligent Robots and Systems},
  2007, pp. 3074--3081.

\bibitem{berenson2009manipulation}
D.~Berenson, S.~S. Srinivasa, D.~Ferguson, A.~Collet, and J.~J. Kuffner,
  ``Manipulation planning with workspace goal regions,'' in \emph{Proceedings
  of IEEE International Conference on Robotics and Automation}, 2009, pp.
  618--624.

\bibitem{Park:2012:ICAPS}
C.~Park, J.~Pan, and D.~Manocha, ``{ITOMP}: Incremental trajectory optimization
  for real-time replanning in dynamic environments,'' in \emph{Proceedings of
  International Conference on Automated Planning and Scheduling}, 2012.

\bibitem{ferguson2006replanning}
D.~Ferguson, N.~Kalra, and A.~Stentz, ``Replanning with rrts,'' in
  \emph{Proceedings 2006 IEEE International Conference on Robotics and
  Automation, 2006. ICRA 2006.}\hskip 1em plus 0.5em minus 0.4em\relax IEEE,
  2006, pp. 1243--1248.

\bibitem{hauser2012responsiveness}
K.~Hauser, ``On responsiveness, safety, and completeness in real-time motion
  planning,'' \emph{Autonomous Robots}, vol.~32, no.~1, pp. 35--48, 2012.

\bibitem{koenig2005fast}
S.~Koenig and M.~Likhachev, ``Fast replanning for navigation in unknown
  terrain,'' \emph{IEEE Transactions on Robotics}, vol.~21, no.~3, pp.
  354--363, 2005.

\bibitem{Kurniawati08sarsop:efficient}
H.~Kurniawati, D.~Hsu, and W.~S. Lee, ``Sarsop: Efficient point-based pomdp
  planning by approximating optimally reachable belief spaces,'' in \emph{In
  Proc. Robotics: Science and Systems}, 2008.

\bibitem{levine2014learning}
S.~Levine and P.~Abbeel, ``Learning neural network policies with guided policy
  search under unknown dynamics,'' in \emph{Advances in Neural Information
  Processing Systems}, 2014, pp. 1071--1079.

\bibitem{park2014high}
C.~Park, J.~Pan, and D.~Manocha, ``High-dof robots in dynamic environments
  using incremental trajectory optimization,'' \emph{International Journal of
  Humanoid Robotics}, vol.~11, no.~02, p. 1441001, 2014.

\bibitem{schulman2016learning}
J.~Schulman, J.~Ho, C.~Lee, and P.~Abbeel, ``Learning from demonstrations
  through the use of non-rigid registration,'' in \emph{Robotics
  Research}.\hskip 1em plus 0.5em minus 0.4em\relax Springer, 2016, pp.
  339--354.

\bibitem{li2015folding}
Y.~Li, Y.~Yue, D.~Xu, E.~Grinspun, and P.~K. Allen, ``Folding deformable
  objects using predictive simulation and trajectory optimization,'' in
  \emph{Intelligent Robots and Systems (IROS), 2015 IEEE/RSJ International
  Conference on}.\hskip 1em plus 0.5em minus 0.4em\relax IEEE, 2015, pp.
  6000--6006.

\bibitem{Chentanez:2009:ISN}
\BIBentryALTinterwordspacing
N.~Chentanez, R.~Alterovitz, D.~Ritchie, L.~Cho, K.~K. Hauser, K.~Goldberg,
  J.~R. Shewchuk, and J.~F. O'Brien, ``Interactive simulation of surgical
  needle insertion and steering,'' in \emph{Proceedings of ACM SIGGRAPH 2009},
  Aug 2009, pp. 88:1--10. [Online]. Available:
  \url{http://graphics.berkeley.edu/papers/Chentanez-ISN-2009-08/}
\BIBentrySTDinterwordspacing

\bibitem{treuille2003keyframe}
A.~Treuille, A.~McNamara, Z.~Popovi{\'c}, and J.~Stam, ``Keyframe control of
  smoke simulations,'' in \emph{ACM Transactions on Graphics (TOG)}, vol.~22,
  no.~3.\hskip 1em plus 0.5em minus 0.4em\relax ACM, 2003, pp. 716--723.

\bibitem{DBLP:journals/corr/PanM16}
\BIBentryALTinterwordspacing
Z.~Pan and D.~Manocha, ``Motion planning for fluid manipulation using
  simplified dynamics,'' \emph{CoRR}, vol. abs/1603.02347, 2016. [Online].
  Available: \url{http://arxiv.org/abs/1603.02347}
\BIBentrySTDinterwordspacing

\bibitem{2016-TOG-deepRL}
X.~B. Peng, G.~Berseth, and M.~van~de Panne, ``Terrain-adaptive locomotion
  skills using deep reinforcement learning,'' \emph{ACM Transactions on
  Graphics (Proc. SIGGRAPH 2016)}, vol.~35, no.~5, 2016, to appear.

\bibitem{bowen2014closed}
C.~Bowen and R.~Alterovitz, ``Closed-loop global motion planning for reactive
  execution of learned tasks,'' in \emph{2014 IEEE/RSJ International Conference
  on Intelligent Robots and Systems}.\hskip 1em plus 0.5em minus 0.4em\relax
  IEEE, 2014, pp. 1754--1760.

\bibitem{yamaguchi2014learning}
A.~Yamaguchi, C.~G. Atkeson, S.~Niekum, and T.~Ogasawara, ``Learning pouring
  skills from demonstration and practice,'' in \emph{2014 IEEE-RAS
  International Conference on Humanoid Robots}.\hskip 1em plus 0.5em minus
  0.4em\relax IEEE, 2014, pp. 908--915.

\bibitem{brandi2014generalizing}
S.~Brandi, O.~Kroemer, and J.~Peters, ``Generalizing pouring actions between
  objects using warped parameters,'' in \emph{2014 IEEE-RAS International
  Conference on Humanoid Robots}.\hskip 1em plus 0.5em minus 0.4em\relax IEEE,
  2014, pp. 616--621.

\bibitem{wang2009physically}
H.~Wang, M.~Liao, Q.~Zhang, R.~Yang, and G.~Turk, ``Physically guided liquid
  surface modeling from videos,'' in \emph{ACM Transactions on Graphics (TOG)},
  vol.~28, no.~3.\hskip 1em plus 0.5em minus 0.4em\relax ACM, 2009, p.~90.

\bibitem{gregson2012stochastic}
J.~Gregson, M.~Krimerman, M.~B. Hullin, and W.~Heidrich, ``Stochastic
  tomography and its applications in 3d imaging of mixing fluids.'' \emph{ACM
  Trans. Graph.}, vol.~31, no.~4, pp. 52--1, 2012.

\bibitem{AISTATS2011_RossGB11}
\BIBentryALTinterwordspacing
S.~Ross, G.~J. Gordon, and D.~Bagnell, ``A reduction of imitation learning and
  structured prediction to no-regret online learning,'' in \emph{Proceedings of
  the Fourteenth International Conference on Artificial Intelligence and
  Statistics (AISTATS-11)}, G.~J. Gordon and D.~B. Dunson, Eds., vol.~15.\hskip
  1em plus 0.5em minus 0.4em\relax Journal of Machine Learning Research -
  Workshop and Conference Proceedings, 2011, pp. 627--635. [Online]. Available:
  \url{http://www.jmlr.org/proceedings/papers/v15/ross11a/ross11a.pdf}
\BIBentrySTDinterwordspacing

\bibitem{2014-cgps}
S.~Levine and V.~Koltun, ``Learning complex neural network policies with
  trajectory optimization,'' in \emph{ICML '14: Proceedings of the 31st
  International Conference on Machine Learning}, 2014.

\bibitem{hansen2016cma}
N.~Hansen, ``The cma evolution strategy: A tutorial,'' \emph{arXiv preprint
  arXiv:1604.00772}, 2016.

\bibitem{karaman2011sampling}
S.~Karaman and E.~Frazzoli, ``Sampling-based algorithms for optimal motion
  planning,'' \emph{The International Journal of Robotics Research}, vol.~30,
  no.~7, pp. 846--894, 2011.

\bibitem{punjani2015deep}
A.~Punjani and P.~Abbeel, ``Deep learning helicopter dynamics models,'' in
  \emph{2015 IEEE International Conference on Robotics and Automation
  (ICRA)}.\hskip 1em plus 0.5em minus 0.4em\relax IEEE, 2015, pp. 3223--3230.

\end{thebibliography}

\begin{changedBlk}
\section{APPENDIX: FLUID SIMULATOR}\label{appen:sph}
We briefly introduce the discrete time-integration function $f$ of the Navier-Stokes equation. Using a particle-based spatial discretization as illustrated in \prettyref{Fig:discrete} (a), we store on each particle its position $p$ and velocity $u$, giving a velocity field. The velocity field evolves in time according to the following equation:
\begin{eqnarray*}
\label{eq:NS}
&&\FDM{u}{t}=\mu\nabla\cdot(\nabla u+\nabla u^T)+g-\nabla p\quad\nabla\cdot u=0,
\end{eqnarray*}
and the position evolves according to the velocity field as $\FDM{p}{t}=u$. Finally, the rigid bodies are taken as boundary conditions by enforcing $\E{L}\cap\E{S}=\E{L}\cap\E{T}=\varnothing$ where $\mu,g,p$ are the liquid viscosity coefficient, gravity, and pressure, respectively. The advection term $\FDM{u}{t}$ allows continuous shape deformation, which makes it very challenging to model, perceive, or predict the shape of the liquid body.
\end{changedBlk}
\end{document}